\documentclass{article}

\usepackage{amsmath,amsfonts,bm}

\def\eqref#1{equation~\ref{#1}}

\def\1{\bm{1}}

\DeclareMathAlphabet{\mathsfit}{\encodingdefault}{\sfdefault}{m}{sl}
\SetMathAlphabet{\mathsfit}{bold}{\encodingdefault}{\sfdefault}{bx}{n}

\usepackage[dvipsnames]{xcolor}
\definecolor{MidnightBlue}{RGB}{25, 25, 112}
\definecolor{RoyalBlue}{RGB}{65, 105, 225} 
\usepackage{graphicx}
\usepackage{listings}
\usepackage{booktabs}
\usepackage{multirow}
\usepackage{enumitem}
\usepackage{subcaption}
\usepackage{caption}
\usepackage{framed}
\usepackage{fancybox}
\usepackage{soul}
\usepackage{tcolorbox}
\usepackage{tabularx}
\usepackage{amssymb}
\usepackage{pifont}
\colorlet{punct}{red!60!black}
\definecolor{background}{HTML}{EEEEEE}
\definecolor{delim}{RGB}{20,105,176}
\colorlet{numb}{magenta!60!black}

\usepackage{adjustbox}

\lstdefinelanguage{json}{
    basicstyle=\scriptsize\ttfamily,
    showstringspaces=false,
    breaklines=true,
    frame=single,
    backgroundcolor=\color{white},
    literate=
     *{0}{{{\color{numb}0}}}{1}
      {1}{{{\color{numb}1}}}{1}
      {2}{{{\color{numb}2}}}{1}
      {3}{{{\color{numb}3}}}{1}
      {4}{{{\color{numb}4}}}{1}
      {5}{{{\color{numb}5}}}{1}
      {6}{{{\color{numb}6}}}{1}
      {7}{{{\color{numb}7}}}{1}
      {8}{{{\color{numb}8}}}{1}
      {9}{{{\color{numb}9}}}{1}
      {:}{{{\color{punct}{:}}}}{1}
      {,}{{{\color{punct}{,}}}}{1}
      {\{}{{{\color{delim}{\{}}}}{1}
      {\}}{{{\color{delim}{\}}}}}{1}
      {[}{{{\color{delim}{[}}}}{1}
      {]}{{{\color{delim}{]}}}}{1},
}

\usepackage{iclr2027_conference,times}
\iclrfinalcopy 

\usepackage[utf8]{inputenc} 
\usepackage[T1]{fontenc}    
\usepackage{hyperref}       
\usepackage{url}            
\usepackage{booktabs}       
\usepackage{amsfonts}       
\usepackage{nicefrac}       
\usepackage{microtype}      
\usepackage{xcolor}         

\usepackage{amsmath}
\usepackage{mathtools}
\usepackage{mdframed}
\usepackage{soul}
\usepackage{xcolor}
\usepackage{ragged2e}

\usepackage{float}

\usepackage{algorithm}
\usepackage{algpseudocode}

\sethlcolor{lightgray}
\definecolor{lightgreen}{rgb}{0.56, 0.93, 0.56}
\definecolor{deepblue}{HTML}{003D79}

\title{\capsysname: Breaking the Scaling Ceiling of Speculative Decoding with Parallel Tree Drafting}

\author{
    Lanxiang Hu$^{1}$
    \hspace{1pt}
    Zhaoxiang Feng$^{1}$
    \hspace{1pt}
    Yulun Wu$^{2}$
    \hspace{1pt}
    Haoran Yuan$^{3}$
    \hspace{1pt}
    Yujie Zhao$^{1}$
    \hspace{1pt}
    Yu-Yang Qian$^{4}$
    \And
    \hspace{1pt}
    Bojun Wang$^{5}$
    \hspace{1pt}
    Peng Zhao$^{4}$
    \hspace{1pt}
    Daxin Jiang$^{5}$
    \hspace{1pt}
    Yibo Zhu$^{5}$
    \hspace{1pt}
    Tajana Rosing$^{1}$
    \hspace{1pt}
    Hao Zhang$^{1}$\\
    $^1$UC San Diego
    \hspace{1pt}
    $^2$ Zhejiang University
    \hspace{1pt}
    $^3$UIUC
    \hspace{1pt}
    $^4$Nanjing University
    \hspace{1pt}
    $^5$StepFun
}

\newcommand{\capsysname}{\textsc{JetSpec}}
\newcommand{\capsysnamespace}{\textsc{JetSpec }}
\newcommand{\sysname}{\text{JetSpec }}
\newcommand{\sysnamenospace}{\text{JetSpec}}

\begin{document}

\maketitle

\begin{abstract}

Speculative decoding (SD) accelerates autoregressive Large Language Models (LLMs) by drafting multiple tokens and verifying them in parallel, but it faces a scaling limitation: increasing the draft budget improves speed only when acceptance remains high and drafting overhead stays low. This ceiling has been difficult to break because prior head-based SD methods face a causality-efficiency dilemma. Autoregressive drafters produce path-conditioned candidates that are effective for tree speculative decoding with higher acceptance length, but their drafting cost grows with tree depth. Bidirectional block-diffusion drafters generate all positions in one pass, but their branch-agnostic marginals can form individually plausible yet mutually inconsistent trees, wasting budget and reducing acceptance. We propose \capsysname, a head-based SD framework that combines one-forward drafting efficiency with causal information flow across draft representations. \sysname trains a causal parallel draft head using fused hidden states from the frozen target model, producing causally coupled draft distributions whose scores better align with the target model's autoregressive factorization. This enables \sysname to convert larger draft budgets into longer accepted prefixes and higher end-to-end speedup. Across math, coding, and chat benchmarks on dense and MoE Qwen3 models, \sysname consistently outperforms bidirectional-head and tree-based SD baselines. On H100 GPUs, \sysname achieves up to $9.64\times$ speedup on MATH-500 and $4.58\times$ on open-ended conversational workloads, with further latency gains demonstrated through vLLM integration under realistic serving loads. Our code and models are available at: \textcolor{blue}{https://github.com/hao-ai-lab/JetSpec}.

\end{abstract}

\section{Introduction}

Modern autoregressive (AR) Large Language Models (LLMs) operate under a simple yet costly bottleneck: decoding remains largely serial. This sequential generation process makes latency a major challenge when they are deployed in applications such as math~\citep{wei2022chain, zhu2025chain}, coding~\citep{Deng2025SWEBenchPro, jain2024livecodebench} and other agentic reasoning tasks~\citep{patil2025bfcl,yao2025taubench,lu2025toolsandbox}, where models often perform long generations.

Existing approaches address this challenge by either incorporating multi-token prediction (MTP) during pre-training or mid-training~\citep{gloeckle2024mtp,deepseek2024v3,deepseek2026v4,huang2026step35flash,xiao2026mimov2flash}, or post-training methods that adapt pretrained LLMs into diffusion LLMs (dLLMs) at the costs of varying degrees of performance degradation~\citep{zhao2025d1,wu2025fastdllmv2}. In contrast, applying speculative decoding (SD) at inference time to reduce generation latency while preserving quality~\citep{leviathan2023speculative, chen2023accelerating}.

Among these techniques, SD stands out for its low adaptation cost and simplicity. On one hand, prior work shows that SD performance can be substantially improved through lightweight draft-model alignment, oftentimes less than 1B training tokens~\citep{liu2024online, zhou2024distillspec, goel2024direct}. On the other hand, head-based methods such as Medusa and EAGLE reduce deployment complexity by avoiding the conventional separate draft-target model setup~\citep{cai2024medusa,li2024eagle}. More recent head-based methods such as EAGLE-3~\citep{li2025eagle3} and DFlash~\citep{chen2026dflash} continue to improve drafting quality and reducing drafting cost for better end-to-end acceleration. Despite these advances, head-based SD still faces a causality-efficiency dilemma. Autoregressive drafters such as EAGLE produce high-quality path-conditioned candidates but require sequential draft passes as tree depth grows. Bidirectional block-diffusion drafters such as DFlash produces drafts in one pass, but their branch-agnostic marginals can form individually plausible yet mutually inconsistent trees, wasting budget and reducing acceptance. 

We propose \capsysname, a SD framework designed to break this causality--efficiency dilemma: \sysname trains a causal parallel draft head to predict distributions for multiple draft depths in one forward pass while enforcing left-to-right information flow across draft-position hidden states. This produces causal depth-wise marginals that better reflect autoregressive continuation structure, enabling tree acceptance to scale more effectively with draft budget.

Empirically, \sysname demonstrates significant efficiency gains. In the low-token-budget regime with 16 draft tokens, \sysname achieves competitive end-to-end speedup.
In the high-token-budget regime with 256 draft tokens, \sysname achieves strong gains, reaching $\tau=10.7$ and $9.5\times$ speedup on MATH-500, and remaining effective on MT-Bench with $\tau=5.9$ and over $4\times$ speedup. We further integrate \sysname into industry-grade serving engine and evaluate it across batch sizes, where it consistently outperforms baselines under small to moderate serving loads, with especially strong gains on higher-end GPUs such as H100 and B200. These results highlight the practical potential of \sysname for reducing decoding latency in serving.


\begin{figure}[t]
    \centering
    \includegraphics[width=0.95\textwidth]{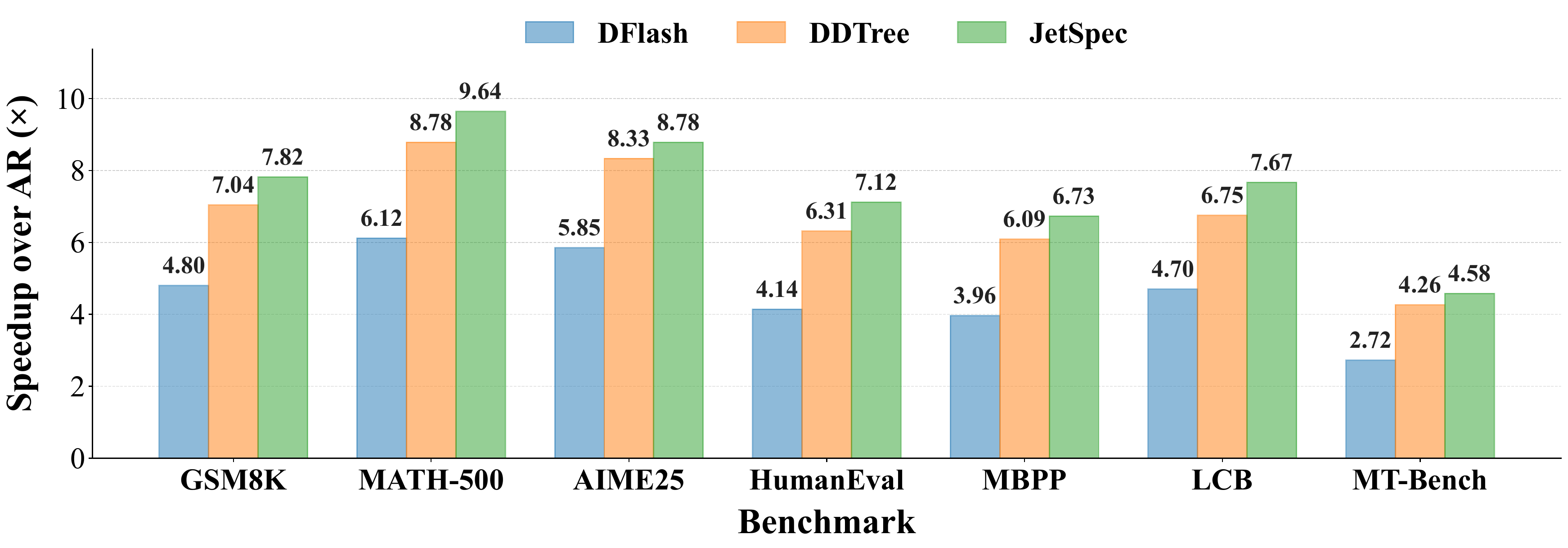}
    \caption{
    End-to-end decoding speedup over standard autoregressive decoding on H100 GPUs across math, coding, and chat benchmarks. DFlash denotes the original block-parallel drafting method, DDTree is tree-based variant of DFlash, and \sysname denotes our method. Both employ a tree budget of 256 tokens using Algorithm~\ref{alg:tree_construction}.
    }
    \label{fig:end_to_end_speedup}
\end{figure}

In a nutshell, our work makes the following contributions:

\begin{itemize}
    \item We propose \capsysname, a paradigm designed to jointly optimize drafting cost and acceptance rate for speculative decoding. \sysname trains a causal prediction head with parallel tree-drafting capability, enabling low-cost draft generation while preserving causal dependencies across draft-depth representations.

    \item We develop and evaluate tree-drafting algorithms that allow speculative decoding to better leverage additional decoding compute. By scaling the draft token budget, \sysname achieves up to $\tau=10.7$ and more than $9.5\times$ end-to-end speedup.

    \item We integrate \sysname into an industry-grade serving engine and evaluate it under realistic serving scenarios. \sysname consistently outperforms baselines under small to moderate serving loads, highlighting its practical potential for reducing latency in real-world deployments.
\end{itemize}

\section{Related Work}

\paragraph{Speculative Decoding.}
Speculative decoding accelerates autoregressive generation by drafting tokens and verifying them with the target model~\citep{leviathan2023speculative,chen2023accelerating}. Its speedup depends on acceptance length and drafting cost. Prior work improves these factors via draft-target alignment~\citep{zhou2024distillspec,liu2024online,goel2024direct}, head-based drafters such as Medusa~\citep{cai2024medusa} and EAGLE~\citep{li2024eagle,li2025eagle3}, or draft-free retrieval methods such as Prompt Lookup Decoding~\citep{saxena2023promptlookup}, REST~\citep{he2024rest}, PLD+~\citep{somasundaram2025pldplus}, SAM Decoding~\citep{hu2025samdecoding}, and SuffixDecoding~\citep{oliaro2024suffixdecoding}. While these methods reduce overhead, they often rely on semantic overlaps for retrieval, or repeated patterns. In contrast, \sysname targets low-cost parallel drafting with a learned causal head that preserves branch-wise dependencies for high-acceptance tree verification.


\paragraph{Drafter Alignment.}
Previous works train draft models or draft heads to better match the target model. Multi-token prediction objectives supervise future-token predictions during pretraining or midtraining~\citep{gloeckle2024mtp,deepseek2024v3}, while post-training approaches use supervised fine-tuning, offline distillation, online distillation, or direct acceptance-rate optimization to improve agreement between draft and target distributions~\citep{zhou2024distillspec,liu2024online,goel2024direct}. These methods primarily improve the alignment term that controls acceptance rate. \sysname is complementary: it trains the drafter not only to match the target model, but also to preserve directional structure for budgeted tree construction, so tree expansion uses causally coupled draft distributions rather than independent position marginals.


\paragraph{Parallel Decoding and Drafting.}
Parallel decoding methods reduce latency by relaxing strict one-token-at-a-time generation. Self-speculative decoding drafts with early target-model layers and verifies with later layers~\citep{elhoushi2024layerskip}. Diffusion LLMs refine multiple tokens per iteration~\citep{wu2025fastdllmv2,qian2026d3llm}. Jacobi-style methods preserve causality by parallel fixed-point updates, with recent consistency distillation works improving convergence to autoregressive outputs~\citep{fu2024lookahead,kou2024cllm,hu2025jacobiforcing}. These methods increase tokens committed per forward, but are typically applied to the target model itself. DFlash generates multiple draft tokens in one block-parallel pass~\citep{chen2026dflash}; DDTree further constructs trees from its bidirectional per-position distributions~\citep{ringel2026ddtree}, while Domino adds a lightweight sequential causal-correction head on top of a parallel drafter to condition later predictions on sampled earlier tokens~\citep{huang2026domino}; see Appendix~\ref{app:parallel_tree_drafting_algorithm} for a detailed comparison. In contrast, \sysname introduces causal information flow directly within the parallel draft head, producing causally coupled depth-wise distributions for tree construction in a single forward pass.
\section{\capsysname}
\label{sec:method}

\subsection{Background}


\paragraph{Scaling Bottlenecks in Speculative Decoding.}

\begin{figure}[t]
    \centering
    \begin{subfigure}[t]{0.48\linewidth}
        \centering
        \includegraphics[width=\linewidth]{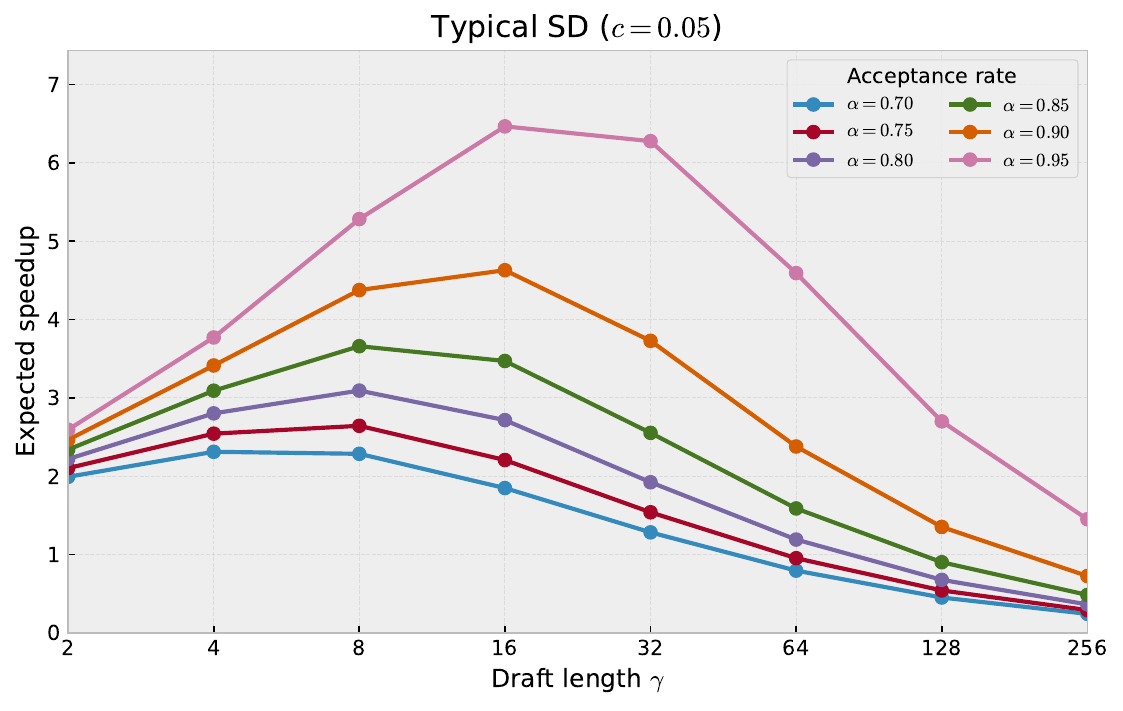}
        \caption{Typical SD ($c=0.05$).}
        \label{fig:typical_sd_alpha_sweep}
    \end{subfigure}
    \hfill
    \begin{subfigure}[t]{0.48\linewidth}
        \centering
        \includegraphics[width=\linewidth]{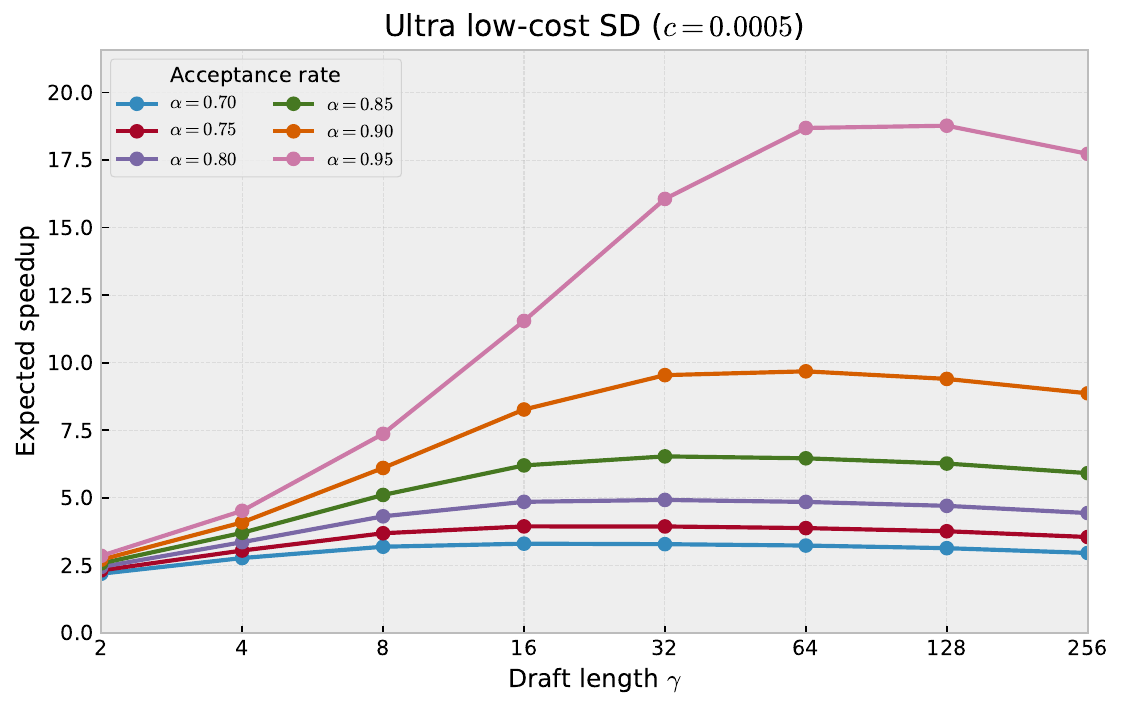}
        \caption{Ultra low-cost SD ($c=0.0005$).}
        \label{fig:cheap_sd_alpha_sweep}
    \end{subfigure}
    \caption{
    Expected speculative decoding speedup scales as a function of draft length $\gamma$, under different per-token drafting costs $c$ and acceptance rates $\alpha$. Comparing the two panels shows that reducing $c$ substantially improves the scalability of speculative decoding with respect to $\gamma$, and increasing $\alpha$ further amplifies this effect. The results highlight that pushing per-token drafting cost $c$ low and acceptance rate $\alpha$ high is critical for unlocking stronger scaling with longer draft lengths.
    }
    \label{fig:sd_speedup_vs_gamma}
\end{figure}

In speculative decoding (SD), a lightweight draft model $M_q$ proposes $N$ tokens, which are verified in parallel by the target model $M_p$~\citep{leviathan2023speculative}. The target accepts the longest prefix consistent with its predictions and resumes from the first rejected token. Let $\alpha$ denote the average acceptance rate and $c$ the per-step cost of $M_q$ relative to $M_p$. Under the standard i.i.d. acceptance assumption,
\begin{equation}
\mathbb{E}[\#\mathrm{tokens}]
=
\frac{1-\alpha^{N+1}}{1-\alpha},
\qquad
\mathrm{Speedup}
=
\frac{1-\alpha^{N+1}}
{(1-\alpha)(Nc+1)}.
\label{eq:sd_speedup}
\end{equation}
Thus, scaling to longer drafts requires both high acceptance $\alpha$ and low total drafting cost $Nc$: as $N$ grows, even modest acceptance degradation or drafting overhead can offset the gains from parallel verification. Figure~\ref{fig:sd_speedup_vs_gamma} illustrates this tradeoff.

Existing work usually improves acceptance rate and drafting cost \textit{separately}. To increase acceptance quality \(\alpha\), draft-model alignment methods improve agreement between \(M_q\) and \(M_p\)~\citep{zhou2024distillspec, liu2024online, goel2024direct}, while tree SD methods improve verification efficiency by expanding multiple candidate branches in parallel~\citep{miao2024specinfer, li2024eagle2}. To reduce drafting cost \(c\), head-based methods avoid a separately deployed draft model~\citep{cai2024medusa, li2024eagle,li2025eagle3}; more recently, DFlash further reduces overhead through block-diffusion-style drafting, generating multiple draft tokens in a single draft pass~\citep{chen2026dflash} (see Appendix~\ref{app:ptd-ptv}). However, scaling to longer drafts requires simultaneously maintaining high \(\alpha\) and low \(c\), motivating \sysname to jointly improve acceptance rate and drafting efficiency.




\subsection{Causal Parallel Drafting}

\paragraph{Can We Realize Both Parallel Drafting and Tree SD?}
A natural direction to co-optimize both aspects is to combine tree drafting for higher acceptance with parallel drafting for lower cost. However, existing parallel drafting methods such as DFlash can break the causal structure needed for effective tree construction: future-position marginals are produced without causal dependence on earlier draft positions, so high-probability marginals can compose into mutually inconsistent branches. \sysname addresses this limitation with a causal parallel draft head that couples draft-depth representations left-to-right while preserving the efficiency of one-pass parallel prediction.


\paragraph{Why Causality Matters.}
\label{par:why_causality}

A tree inherently encodes a causal structure. Consider a draft tree rooted at the original prefix \(x\), where each node $v$ corresponds to a candidate token \(y_v\). Let $\pi(v)=(y_{v_1},\ldots,y_v)$ denote the path from the root to $v$, and $\pi_{<v}$ its ancestor tokens. Ideally, the distribution at each node should reflect its corresponding branch prefix. In contrast, a bidirectional parallel drafter allows each position to attend to representations from both earlier and later draft positions. It creates a mismatch between bidirectional draft representations and the causal structure required for tree construction.

\paragraph{Architecture.}
\sysname adopts a head-based drafting architecture, where a draft head can reuse intermediate features from the target model. Specifically, DFlash shows that a lightweight block-parallel draft head can reuse target-model hidden states and inject fused target-context features into the draft layers' KV cache for strong target-model guidance~\citep{chen2026dflash}. Building on this design, \sysname introduces a causal parallel decoding head for tree drafting.

Given a prefix $x$ with fused target-model features $h_x^{o}$, the draft head predicts logits for multiple future positions in parallel. Following standard speculative decoding notation, the target model defines an autoregressive distribution over a length-$k$ continuation:

\begin{equation}
p(y_{1:k}\mid x)
=
\prod_{i=1}^{k} p(y_i \mid x, y_{<i}).
\label{eq:ar_distribution}
\end{equation}


To introduce analogous directional dependencies into parallel drafting, we apply a causal attention mask across draft positions. Position $i$ may attend to the prefix features and earlier draft-position hidden states, but not to later positions. The mask and the masked attention at position $i$ are


\begin{equation}
B_{i,j} = 
\begin{cases} 0, & j \le i, \\
-\infty, & j > i, 
\end{cases},
\qquad 
\mathrm{Attn}(Q_i,K,V)=
\mathrm{softmax}
\left(
\frac{Q_iK^\top}{\sqrt{d}} + B_i
\right)V,
\end{equation}
with prefix features visible to all draft positions. This allows all draft positions to be processed in parallel while enforcing directional information flow from earlier to later positions. Consequently, the distribution $r_i(\cdot\mid x,h_x^o)$ at position $i$ is computed from a hidden representation that depends causally on earlier draft-position representations, rather than being produced independently across positions.

\begin{figure*}[t]
    \centering
    \includegraphics[width=0.975\textwidth]{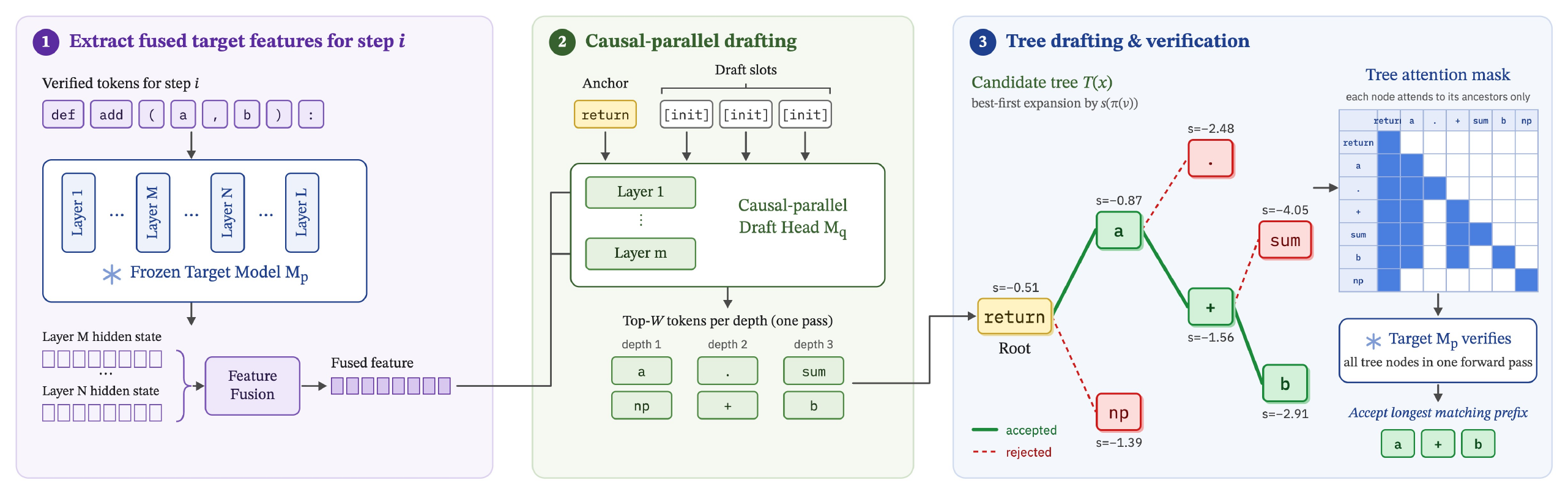}
    \caption{
    \sysname design overview. 
     \sysname extracts fused hidden features from the frozen target model and conditions a causal-parallel draft head to generate high-quality candidate trees in one forward pass.
    }
    \label{fig:design_overall}
\end{figure*}



The causal mask induces directional dependencies across draft-position representations. Let \(h_i^q\) denote the draft-head hidden state at position \(i\). Then
\begin{equation}
h_i^q
=
M_q\!\left(x,h_x^{o},h_{<i}^q\right),
\qquad
q_i
=
\mathrm{softmax}\!\left(W h_i^q\right),
\end{equation}
where $h_{<i}^q$ denotes earlier draft-position representations and $q_i$ the token distribution at position $i$. Although all positions are predicted in parallel, each position causally incorporates earlier representations, introducing directional dependencies for tree drafting.

\subsection{Training}

\paragraph{Data Preparation.}
We train the draft head on target-aligned sequences, obtained either from the training corpus, as in multi-token prediction~\citep{gloeckle2024mtp,deepseek2024v3}, or from 
continuations regenerated by the target model~\citep{agarwal2024gkd,gu2024minillm,liang2026cd4lm}. Given a sequence \((x,y_{1:k})\), we sample anchor positions and, at each anchor, construct a length-$N$ training block of $N$ consecutive future positions. The draft head predicts these $N$ positions in parallel under a block-causal mask, matching the per-position teacher logits produced by the target model on the same ground-truth prefix.

\paragraph{Distillation Loss.}
We train the draft head with a soft-label distillation objective so that the draft distribution preserves the target model's relative preferences across multiple plausible continuations. Among distillation objectives, forward KL encourages the draft distribution to cover the target model's probability mass, preserving soft-label information beyond the top prediction. Reverse KL, in contrast, is more mode-seeking and can concentrate probability on high-confidence teacher modes~\citep{gu2024minillm,agarwal2024gkd}. We find this matters in practice (Section~\ref{sec:experiments}): reverse KL underperforms substantially, while forward KL slightly outperforms hard-label SFT. We therefore adopt forward KL by default, and present the loss in this form below.

For each active draft position $m$, let $z_q^{(m)}$ and $z_p^{(m)}$ denote the draft and target logits over the vocabulary $\mathcal{V}$. We define temperature-normalized distributions
$\tilde{q}^{(m)} = \mathrm{softmax}(z_q^{(m)}/T_{\mathrm{KD}})$
and
$\tilde{p}^{(m)} = \mathrm{softmax}(z_p^{(m)}/T_{\mathrm{KD}})$, and minimize the forward KL
\begin{equation}
    \mathcal{L}_{\mathrm{FKL}}^{(m)}
    =
    D_{\mathrm{KL}}
    \left(
        \tilde{p}^{(m)}
        \,\middle\|\,
        \tilde{q}^{(m)}
    \right).
\end{equation}
The final objective is computed over all active draft-token positions induced from the training sequence and normalized by the active-position mask (see Appendix~\ref{app:training_details} for details):
\begin{equation}
    \mathcal{L}_{\mathrm{train}}
    =
    T_{\mathrm{KD}}^2
    \frac{
        \sum_m w_m
        \mathcal{L}_{\mathrm{FKL}}^{(m)}
    }{
        \sum_m w_m
    },
\end{equation}
where $m$ indexes draft-token positions within sampled blocks, excluding non-loss and anchor-token positions, and $T_{\mathrm{KD}}>0$ is the distillation temperature. This objective preserves the target model's soft-label preferences and provides teacher-aligned candidate distributions for tree expansion.

\subsection{Drafting and Verification}


\paragraph{Parallel Tree Drafting.}
An overview of the design is shown in Figure~\ref{fig:design_overall}. At every decoding step $i$, the causal draft head conditions on the hidden states $h_x^{o}$ of tokens that were verified and accepted in the previous step, and constructs a candidate tree with maximum draft depth $N$, branching width $W$, and total node budget $B$. With one draft-head forward pass, we obtain logits for all draft depths and extract the top-\(W\) candidate tokens at each depth. Tree expansion is then driven by a branch scoring function \(s(\pi(v))\), whose form can vary across implementations; we ablate different choices in Section~\ref{sec:experiment_ablation_studies}. By default, we use accumulated draft log-probability for scoring:
\begin{equation}
    s(\pi(v))
    =
    \sum_{u\in \pi(v)}
    \log r_{d(u)} \left( y_u,\mid x, h_x^o \right),
\label{eq:accum_logp}
\end{equation}
where where $d(u)$ is node depth and $r_d$ is the distribution produced at depth $d$ by the one-pass causal head. We then repeatedly pop the highest-scoring expandable node, expand it with up to \(W\) children, and reinsert the children with updated scores until the budget is filled or no expandable node remains. This yields a candidate tree $\mathcal{T}(x) = \{\pi(v) \mid v \in \mathcal{V}_{\mathcal{T}}\}$, where each path \(\pi(v)\) is a speculative continuation; details are given in Algorithm~\ref{alg:tree_construction}.

\paragraph{Tree Verification.}
After constructing the candidate tree \(\mathcal{T}(x)\), the target model \(M_p\) verifies all tree nodes in parallel using tree attention. For any candidate branch \(\pi(v)=y_{1:k}\), the target distribution factorizes autoregressively in compliance with Equation~\ref{eq:ar_distribution}.

The verifier then applies the speculative decoding acceptance rule along each candidate branch. For each draft token \(y_t\), we write the acceptance decision as
\begin{equation}
    A_t
    \sim
    \mathrm{Bernoulli}(\alpha_t),
    \qquad
    \alpha_t
    =
    \alpha\!\left(
        y_t;\,
        q(\cdot\mid x,y_{<t}),
        p(\cdot\mid x,y_{<t})
    \right),
\end{equation}
where \(q\) is the draft distribution induced by \(M_q\), and \(\alpha(\cdot)\) denotes the speculative verification rule. In standard non-greedy speculative sampling, this acceptance rule uses rejection sampling:
\begin{equation}
    \alpha_t
    =
    \min\!\left(
        1,\,
        \frac{
            p(y_t\mid x,y_{<t})
        }{
            q(y_t\mid x,y_{<t})
        }
    \right),
\label{eq:nongreedy_specdec}
\end{equation}
with a target-model correction token sampled when rejection occurs~\citep{leviathan2023speculative}. The accepted prefix length is $a = \max \left\{ r \le k: A_t = 1,\ \forall t \le r \right\}$. In the greedy setting, \(A_t\) becomes deterministic: a draft token is accepted if it matches the target model's next-token prediction.

\section{Experiments}
\label{sec:experiments}


\subsection{Experiment Setup}

\paragraph{Models and Datasets.}
We evaluate \sysname on Qwen3-8B and Qwen3-30B-A3B, covering both dense and MoE target models~\citep{yang2025qwen3}. Qwen3 supports both thinking and non-thinking modes, and we use the non-thinking mode throughout our main evaluation for efficient decoding. For training data, we curate 780K examples from the Nemotron Post-Training Dataset V2~\citep{nvidia2025nemotronposttrainingv2}, including all available coding and math splits, random samples from STEM and chat splits, and 20K additional examples from CodeAlpaca~\citep{chaudhary2023codealpaca}. For regenerated training sequences, we apply the corresponding chat template to each data type and continue generation with the target model. We evaluate on math benchmarks including GSM8K~\citep{cobbe2021gsm8k}, MATH-500~\citep{hendrycks2021math}, and AIME25; coding benchmarks including HumanEval~\citep{chen2021codex}, MBPP~\citep{austin2021program}, and LiveCodeBench~\citep{jain2024livecodebench}; and open-ended conversational tasks including MT-Bench~\citep{zheng2023judging}. All ablation studies are conducted on the math split of Nemotron Post-Training Dataset V2.

\paragraph{Baselines and Training Settings.}
We compare against EAGLE-3 and DFlash, two strong head-based speculative decoding baselines~\citep{li2025eagle3,chen2026dflash}. EAGLE-3 uses multi-layer feature fusion for direct token prediction, while DFlash uses a bidirectional block-parallel draft head to generate multiple draft tokens in one pass. DDTree~\citep{ringel2026ddtree} constructs draft trees from DFlash's per-position block-diffusion distributions using best-first tree expansion, but uses a pure block-diffusion head trained without loss weighting~\citep{ringel2026ddtree}. For fair comparison, we train \sysname and all baselines on the same data mixture using a learning rate of \(3\times10^{-4}\) on 8 H100 GPUs. We also ablate DFlash-style loss weighting in Section~\ref{sec:experiment_ablation_studies}; additional training details are provided in Appendix~\ref{app:training_details}.


\paragraph{Implementation Details.}
For offline inference evaluation, we run experiments on 8 H100 or B200 GPUs. Unless otherwise specified, results in Table~\ref{tab:main_results_tree_budget_sweep} use optimized Triton kernels for decoding. For serving-engine evaluation, we integrate \sysname into vLLM (Appendix~\ref{app:vllm_implementation_details}), part of which involves tree verification with a custom SM90 paged FlashAttention kernel using NVIDIA CuTe DSL, extending the paged attention forward path with shared-memory tree-mask staging and tree-tail masking.


\subsection{Results} 
\label{sec:experiment_results}

We evaluate \sysname under both greedy and non-greedy decoding settings in compliance with Equation~\ref{eq:nongreedy_specdec}. Evaluation results and comparisons with baselines are summarized in Table~\ref{tab:main_results_linear_budget_sweep} and Table~\ref{tab:main_results_tree_budget_sweep}, and it shows \sysname consistently outperforms tree-based baselines across all benchmarks.


\paragraph{Low-budget Regime.}
Compared with the autoregressive drafting baseline EAGLE-3, \sysname achieves substantially higher speedup at every reported budget by avoiding EAGLE-3's higher sequential drafting overhead. At budget 16, DFlash and \sysname perform nearly the same, suggesting that a short linear draft is already sufficient to cover many high-probability continuations. However, when increasing the budget to 32, \sysname brings slight improvement, while DFlash often saturates or degrades. This indicates that \sysname converts the additional draft budget into accepted tokens more effectively.


\begin{table*}[t]
\centering
\small
\caption{
Low-budget regime comparison of \sysname and baselines trained with the same Qwen3-8B model and data recipe. We report results on math, coding, and chat benchmarks using non-thinking mode with a 3072 max tokens. We report end-to-end decoding speedup over standard AR decoding and average accepted length $\tau$ on H100 GPU.
}
\label{tab:main_results_linear_budget_sweep}
\setlength{\tabcolsep}{3.2pt}
\begin{adjustbox}{max width=\textwidth}
\begin{tabular}{lccccccccccccccc}
\toprule
\multirow{2}{*}{Method}
& \multirow{2}{*}{Budget}
& \multicolumn{2}{c}{GSM8K}
& \multicolumn{2}{c}{MATH-500}
& \multicolumn{2}{c}{AIME25}
& \multicolumn{2}{c}{HumanEval}
& \multicolumn{2}{c}{MBPP}
& \multicolumn{2}{c}{LCB}
& \multicolumn{2}{c}{MT-Bench} \\
\cmidrule(lr){3-4}
\cmidrule(lr){5-6}
\cmidrule(lr){7-8}
\cmidrule(lr){9-10}
\cmidrule(lr){11-12}
\cmidrule(lr){13-14}
\cmidrule(lr){15-16}
& 
& Speedup & $\tau$
& Speedup & $\tau$
& Speedup & $\tau$
& Speedup & $\tau$
& Speedup & $\tau$
& Speedup & $\tau$
& Speedup & $\tau$ \\
\midrule

\multicolumn{16}{c}{Temperature = 0} \\
\midrule
\multirow{2}{*}{EAGLE-3}
& 16  & 2.24 & 3.78 & 2.10 & 3.61 & 2.08 & 3.55 & 2.18 & 3.72 & 1.95 & 3.31 & 1.82 & 3.19 & 1.91 & 3.40 \\
& 32  & 2.39 & 4.03 & 2.23 & 3.87 & 2.22 & 3.80 & 2.34 & 4.00 & 2.09 & 3.60 & 1.96 & 3.42 & 2.04 & 3.62 \\
\midrule
\multirow{2}{*}{DFlash}
& 16  & 4.80 & 6.01 & 6.12 & 7.83 & \textbf{5.85} & \textbf{7.34} & 4.14 & 5.18 & 3.96 & 4.98 & 4.70 & 6.16 & \textbf{2.72} & 4.03 \\
& 32  & 4.21 & 5.27 & 5.39 & 6.89 & 5.15 & 6.38 & 3.63 & 4.53 & 3.51 & 4.39 & 3.99 & 5.23 & 2.48 & 3.61 \\
\midrule
\multirow{2}{*}{\sysname}
& 16  & 4.80 & 6.00 & 6.06 & 7.75 & 5.78 & 7.23 & 4.26 & 5.32 & 3.97 & 4.96 & 4.77 & 6.24 & 2.68 & 3.98 \\
& 32  & \textbf{4.89} & \textbf{6.14} & \textbf{6.35} & \textbf{8.23} & 5.75 & 7.25 & \textbf{4.29} & \textbf{5.35} & \textbf{4.03} & \textbf{5.03} & \textbf{4.86} & \textbf{6.48} & 2.67 & \textbf{4.09} \\

\midrule
\multicolumn{16}{c}{Temperature = 1} \\
\midrule
\multirow{2}{*}{EAGLE-3}
& 16  & 2.16 & 3.65 & 2.01 & 3.48 & 1.89 & 3.24 & 2.02 & 3.58 & 1.89 & 3.22 & 1.81 & 3.09 & 1.85 & 3.24 \\
& 32  & 2.26 & 3.90 & 2.12 & 3.74 & 2.01 & 3.49 & 2.19 & 3.86 & 2.02 & 3.50 & 1.91 & 3.31 & 1.90 & 3.42 \\
\midrule
\multirow{2}{*}{DFlash}
& 16  & 4.32 & 5.44 & 4.87 & 6.30 & 3.55 & 4.69 & 3.56 & 4.45 & 3.52 & 4.43 & 4.38 & 5.70 & \textbf{2.41} & 3.52 \\
& 32  & 3.85 & 4.84 & 4.29 & 5.55 & 3.44 & 4.48 & 3.22 & 4.00 & 3.17 & 3.97 & 3.74 & 4.81 & 2.23 & 3.20 \\
\midrule
\multirow{2}{*}{\sysname}
& 16  & 4.32 & 5.44 & 4.77 & 6.18 & 3.48 & 4.54 & 3.70 & 4.63 & 3.52 & 4.41 & 4.41 & 5.74 & \textbf{2.41} & 3.53 \\
& 32  & \textbf{4.40} & \textbf{5.58} & \textbf{4.91} & \textbf{6.44} & \textbf{3.64} & \textbf{4.76} & \textbf{3.72} & \textbf{4.66} & \textbf{3.60} & \textbf{4.51} & \textbf{4.53} & \textbf{5.93} & 2.40 & \textbf{3.54} \\
\bottomrule
\end{tabular}
\end{adjustbox}
\end{table*}

\paragraph{High-budget Regime.}
In Table~\ref{tab:main_results_tree_budget_sweep}, as the draft budget increases, \sysname shows a larger gain, e.g. improving from roughly $4$--$6\times$ to $7$--$10\times$ speedup on math and coding benchmarks. Whereas DDTree scales more modestly to up to $9\times$, showing causal tree drafting yields higher acceptance than branch-agnostic tree construction. And detailed qualitative example in Appendix~\ref{app:tree_quality_top5} on how causality helps. Under temperature $1$, \sysname remains effective, showing that the benefit of causal tree drafting remains robust under nongreedy decoding.

\begin{table*}[t]
\centering
\small

\caption{
High-budget comparison on Qwen3-8B with at least 64 draft tokens. EAGLE-3 uses tree mode with max depth 8; larger budgets give minimal or worse gains due to training mismatch.
}


 

\label{tab:main_results_tree_budget_sweep}
\setlength{\tabcolsep}{3.2pt}
\begin{adjustbox}{max width=\textwidth}
\begin{tabular}{lccccccccccccccc}
\toprule
\multirow{2}{*}{Method}
& \multirow{2}{*}{Budget}
& \multicolumn{2}{c}{GSM8K}
& \multicolumn{2}{c}{MATH-500}
& \multicolumn{2}{c}{AIME25}
& \multicolumn{2}{c}{HumanEval}
& \multicolumn{2}{c}{MBPP}
& \multicolumn{2}{c}{LCB}
& \multicolumn{2}{c}{MT-Bench} \\
\cmidrule(lr){3-4}
\cmidrule(lr){5-6}
\cmidrule(lr){7-8}
\cmidrule(lr){9-10}
\cmidrule(lr){11-12}
\cmidrule(lr){13-14}
\cmidrule(lr){15-16}
& & Speedup & $\tau$ & Speedup & $\tau$ & Speedup & $\tau$ & Speedup & $\tau$ & Speedup & $\tau$ & Speedup & $\tau$ & Speedup & $\tau$ \\
\midrule
\multicolumn{16}{c}{Temperature = 0} \\
\midrule
\multirow{1}{*}{EAGLE-3}
& 64  & 2.53 & 4.31 & 2.36 & 4.13 & 2.35 & 4.04 & 2.49 & 4.26 & 2.22 & 3.81 & 2.09 & 3.62 & 2.19 & 3.88 \\
\midrule


\multirow{3}{*}{DDTree}
& 64  & 5.63 & 6.18 & 6.51 & 7.16 & 6.40 & 6.96 & 5.08 & 5.57 & 4.99 & 5.49 & 5.47 & 6.06 & 3.74 & 4.51 \\
& 128 & 6.63 & 7.31 & 8.27 & 9.19 & 7.93 & 8.66 & 5.93 & 6.52 & 5.70 & 6.28 & 6.42 & 7.25 & 4.12 & 5.14 \\
& 256 & 7.04 & 7.77 & 8.78 & 9.81 & 8.33 & 9.24 & 6.31 & 6.96 & 6.09 & 6.70 & 6.75 & 7.72 & 4.26 & 5.41 \\
\midrule

\multirow{3}{*}{\sysname}
& 64  & 5.98 & 6.56 & 6.76 & 7.42 & 6.47 & 7.00 & 5.53 & 6.06 & 5.34 & 5.88 & 5.95 & 6.59 & 3.97 & 4.77 \\
& 128 & 7.34 & 8.05 & 8.93 & 9.95 & 8.26 & 9.10 & 6.66 & 7.28 & 6.31 & 6.95 & 7.29 & 8.21 & 4.37 & 5.52 \\
& 256 & \textbf{7.82} & \textbf{8.62} & \textbf{9.64} & \textbf{10.76} & \textbf{8.78} & \textbf{9.82} & \textbf{7.12} & \textbf{7.78} & \textbf{6.73} & \textbf{7.43} & \textbf{7.67} & \textbf{8.79} & \textbf{4.58} & \textbf{5.94} \\


\midrule
\multicolumn{16}{c}{Temperature = 1} \\
\midrule
\multirow{1}{*}{EAGLE-3}
& 64  & 2.35 & 4.17 & 2.23 & 4.00 & 2.11 & 3.73 & 2.35 & 4.13 & 2.13 & 3.70 & 1.99 & 3.49 & 1.96 & 3.63 \\
\midrule


\multirow{3}{*}{DDTree}
& 64  & 5.28 & 5.77 & 5.68 & 6.26 & 4.94 & 5.46 & 4.65 & 5.09 & 4.65 & 5.11 & 5.28 & 5.82 & 3.50 & 4.22 \\
& 128 & 6.11 & 6.72 & 6.79 & 7.60 & 5.40 & 5.98 & 5.29 & 5.76 & 5.22 & 5.75 & 5.99 & 6.71 & 3.66 & 4.59 \\
& 256 & 6.41 & 7.17 & 7.10 & 8.13 & 5.26 & 6.20 & 5.43 & 6.03 & 5.49 & 6.12 & 6.26 & 7.17 & 3.81 & 4.88 \\
\midrule
\multirow{3}{*}{\sysname}
& 64  & 5.63 & 6.19 & 5.97 & 6.60 & 4.95 & 5.48 & 5.02 & 5.52 & 5.00 & 5.51 & 5.76 & 6.38 & 3.63 & 4.37 \\
& 128 & 6.76 & 7.49 & 7.39 & 8.36 & 5.76 & 6.49 & 5.82 & 6.39 & 5.78 & 6.39 & 6.84 & 7.74 & 3.99 & 5.01 \\
& 256 & \textbf{7.16} & \textbf{8.03} & \textbf{7.83} & \textbf{9.01} & \textbf{5.94} & \textbf{7.06} & \textbf{6.19} & \textbf{6.85} & \textbf{6.11} & \textbf{6.83} & \textbf{7.25} & \textbf{8.29} & \textbf{4.06} & \textbf{5.22} \\
\bottomrule
\end{tabular}
\end{adjustbox}
\end{table*}


\subsection{System Performance}
\label{sec:sys_perf}


We evaluate \sysname in an end-to-end serving setting by integrating tree drafting and verification into vLLM. Serving performance depends not only on accepted length, but also on how the tree budget interacts with batching, verification overhead, memory traffic, and GPU occupancy. Larger budgets can reduce the number of target-model verification rounds, but also introduce higher draft-head and tree-verification cost, making the optimal budget load-dependent.



\paragraph{Budget Allocation.}

Table~\ref{tab:vllm_batch_budget_sweep} shows that larger tree budgets are most useful at small batch sizes. On H100 GPU at batch size 1, increasing the budget from 16 to 128 raises throughput from 224.0 to 553.3 TPS and end-to-end speedup from 1.75$\times$ to 4.33$\times$. At batch size 16, a budget of 32 performs best at 3.81$\times$, while budget 128 drops to 2.80$\times$. This illustrates that larger budgets are most beneficial at low load, whereas heavier batching favors more moderate budgets.

\paragraph{Practical implication.}

These results suggest that tree budgets should be selected according to serving load. Large budgets are preferable in low- to moderate-load regimes, where spare GPU capacity can be used to reduce decoding iterations. Under heavier load, smaller or moderate budgets may be more efficient by reducing per-step overhead and preserving batching efficiency. We evaluate static budget policies in this work and leave dynamic serving-time budget scheduling to future work.




\subsection{Ablation Studies}
\label{sec:experiment_ablation_studies}

Unless noted, all ablations in this section are evaluated on $4{\times}$B200 with block size $16$, FlashAttention-2 for the target's normal forward, Triton tree-attention, and the accumulated draft log-probability heuristic (Equation~\ref{eq:accum_logp}
) for tree construction at node budget $255$. Reported average accepted length $\tau$ are independent of hardware under lossless verification.

\subsubsection{Training Settings}


\begin{table}[ht]
\centering
\small
\caption{
Learning-rate ablation with \capsysnamespace and without loss weighting training ($\gamma=0$). See Section~\ref{sec:experiment_diffusion_tree_weakness} for $\gamma$'s definition and ablations. We report speedup and average accepted length $\tau$.
}
\label{tab:ablation_lr}
\setlength{\tabcolsep}{4.5pt}
\begin{tabular}{lcccccccc}
\toprule
\multirow{2}{*}{LR}
& \multicolumn{4}{c}{GSM8K}
& \multicolumn{4}{c}{MATH-500} \\
\cmidrule(lr){2-5}
\cmidrule(lr){6-9}
& \multicolumn{2}{c}{SFT}
& \multicolumn{2}{c}{Forward KL}
& \multicolumn{2}{c}{SFT}
& \multicolumn{2}{c}{Forward KL} \\
\cmidrule(lr){2-3}
\cmidrule(lr){4-5}
\cmidrule(lr){6-7}
\cmidrule(lr){8-9}
& Speedup & $\tau$ & Speedup & $\tau$
& Speedup & $\tau$ & Speedup & $\tau$ \\
\midrule
$5{\times}10^{-5}$ & 4.69 & 5.46 & 4.76 & 5.58 & 7.27 & 8.70 & 7.29 & 8.71 \\
$1{\times}10^{-4}$ & 5.37 & 6.22 & 5.45 & 6.32 & 7.89 & 9.34 & 7.89 & 9.39 \\
$3{\times}10^{-4}$ & 5.79 & 6.78 & 5.92 & 6.87 & 8.30 & 9.80 & 8.29 & 9.81 \\
$6{\times}10^{-4}$ & 5.87 & 6.74 & 5.79 & 6.78 & 8.23 & 9.73 & 8.14 & 9.70 \\
$1{\times}10^{-3}$ & 5.75 & 6.64 & 5.73 & 6.70 & 8.17 & 9.66 & 8.15 & 9.66 \\
\bottomrule
\end{tabular}
\end{table}

\paragraph{Learning Rates.}
Table~\ref{tab:ablation_lr} reports a grid search over learning rates from $5\times 10^{-5}$ to $1\times 10^{-3}$. Very small learning rates underperform and performance plateaus at $3\times 10^{-4}$ (peak speedup $8.30\times$), with $6\times 10^{-4}$ and $1\times 10^{-3}$ within ${\sim}2\%$ of the peak.


\paragraph{Loss.}
Table~\ref{tab:ablation_loss} compares SFT, forward-KL distillation, and reverse-KL distillation across four math benchmarks. SFT and forward-KL distillation match within ${\sim}3\%$ at every dataset, while reverse-KL distillation causes a $36$--$46\%$ relative drop in comparison with forward-KL's, indicating the mode-seeking objective concentrates draft probability mass too aggressively for tree drafting.

\begin{table}[H]
\centering
\small
\caption{Loss-objective ablation with \capsysnamespace at LR $6{\times}10^{-4}$ and $\gamma=0$. We report speedup and average accepted length $\tau$. All cells use checkpoints from a single training pipeline trained on math-only data ($\sim$3 epochs); we report on math benchmarks to keep the comparison in-distribution.}

\label{tab:ablation_loss}
\setlength{\tabcolsep}{4.5pt}
\begin{tabular}{lcccccc}
\toprule
\multirow{2}{*}{Dataset}
& \multicolumn{2}{c}{SFT}
& \multicolumn{2}{c}{Forward-KL Distill}
& \multicolumn{2}{c}{Reverse-KL Distill} \\
\cmidrule(lr){2-3}\cmidrule(lr){4-5}\cmidrule(lr){6-7}
& Speedup & $\tau$ & Speedup & $\tau$ & Speedup & $\tau$ \\
\midrule
GSM8K    & 5.96 &  6.93 & 6.11 &  7.09 & 3.29 & 3.78 \\
MATH-500 & 8.42 &  9.98 & 8.46 & 10.01 & 5.25 & 6.59 \\
AIME25  & 7.51 &  9.51 & 7.56 &  9.40 & 4.76 & 6.00 \\
AIME24  & 8.25 &  9.93 & 8.00 &  9.70 & 5.13 & 6.43 \\
\bottomrule
\end{tabular}
\end{table}


\paragraph{Model Generalizability.}
Table~\ref{tab:ablation_model_generalizability} compares \sysname and DDTree trained on Qwen3-30B-A3B, a MoE target model, with the same 800k data recipe. \sysname maintains a competitively higher speedups across benchmarks, showing that causal tree drafting generalizes beyond dense Qwen3-8B.

\begin{table}[ht]
\centering
\small
\caption{\sysname generalization to MoE architecture. \sysname vs.\ DDTree on Qwen3-30B-A3B, both trained with SFT on the same 800K-example data mixture as our Qwen3-8B main results. Each cell reports speedup / average accepted length $\tau$ at temperature $0$ with tree budget $256$.}
\label{tab:ablation_model_generalizability}
\setlength{\tabcolsep}{4.0pt}
\begin{adjustbox}{max width=\textwidth}
\begin{tabular}{lccccccc}
\toprule
Method
& GSM8K
& MATH-500
& AIME25
& HumanEval
& MBPP
& LCB
& MT-Bench \\
\midrule
DDTree
& 7.26 / 7.93
& 8.61 / 9.49
& 9.01 / 9.71
& 6.18 / 6.76
& 6.39 / 7.06
& 7.40 / 8.31
& 4.26 / 5.35 \\
\sysname
& 7.40 / 8.18
& 9.45 / 10.65
& 9.35 / 10.28
& 6.51 / 7.23
& 6.53 / 7.29
& 7.47 / 8.62
& 4.33 / 5.59 \\
\bottomrule
\end{tabular}
\end{adjustbox}
\end{table}

\paragraph{Training Data.}
Table~\ref{tab:ablation_training_data} studies the effect of using different training data. 
Regenerated target-model sequences provide the strongest performance, confirming that matching the draft head to the target model's own generations is important for maximizing accepted length. In contrast, \sysnamenospace-Corpus trains directly on corpus sequences, it underperforms fully regenerated \sysnamenospace, but still achieves consistent speedups across all benchmarks, suggesting that causal parallel drafting can potentially be incorporated during mid-training or even pretraining.

\begin{table}[th]
\centering
\small
\caption{
Both \sysname and \sysnamenospace-Corpus are trained with SFT on the same 800K-example data mixture. \sysname uses model-regenerated continuations as supervision targets; \sysnamenospace-Corpus uses the original training corpus. Each cell reports speedup / average accepted length $\tau$ at temperature $0$.
}

\label{tab:ablation_training_data}
\setlength{\tabcolsep}{3.5pt}
\begin{adjustbox}{max width=\textwidth}
\begin{tabular}{lccccccc}
\toprule
Method & Budget
& GSM8K
& AIME25
& HumanEval
& MBPP
& LCB
& MT-Bench \\
\midrule
\capsysname
& 16
& 4.80 / 6.00
& 5.78 / 7.23
& 4.26 / 5.32
& 3.97 / 4.96
& 4.77 / 6.24
& 2.68 / 3.98 \\
\capsysname-Corpus
& 16
& 2.06 / 2.54
& 2.41 / 2.99
& 2.11 / 2.61
& 1.93 / 2.39
& 2.75 / 3.45
& 1.54 / 2.00 \\
\midrule
\capsysname
& 64
& 5.98 / 6.56
& 6.47 / 7.00
& 5.53 / 6.06
& 5.34 / 5.88
& 5.95 / 6.59
& 3.97 / 4.77 \\
\capsysname-Corpus
& 64
& 2.57 / 2.78
& 2.70 / 2.89
& 2.74 / 2.97
& 2.60 / 2.83
& 3.43 / 3.64
& 2.28 / 2.40 \\
\midrule
\capsysname
& 256
& 7.82 / 8.62
& 8.78 / 9.82
& 7.12 / 7.78
& 6.73 / 7.43
& 7.67 / 8.79
& 4.58 / 5.94 \\
\capsysname-Corpus
& 256
& 3.36 / 3.65
& 3.66 / 4.06
& 3.53 / 3.82
& 3.27 / 3.58
& 4.42 / 4.86
& 2.63 / 2.98 \\
\bottomrule
\end{tabular}
\end{adjustbox}
\end{table}




\subsubsection{Tree Drafting with Diffusion Head}
\label{sec:experiment_diffusion_tree_weakness}

\begin{table}[th]
\centering
\small
\caption{Architecture and $\gamma$ ablation on MATH-500 with \sysname at LR $3{\times}10^{-4}$ with forward-KL distillation. We report speedup and average accepted length $\tau$.}
\label{tab:ablation_gamma_arch}
\setlength{\tabcolsep}{4.5pt}
\begin{tabular}{lcccccccc}
\toprule
\multirow{2}{*}{Head}
& \multicolumn{2}{c}{$\gamma=0$}
& \multicolumn{2}{c}{$\gamma=3$}
& \multicolumn{2}{c}{$\gamma=7$}
& \multicolumn{2}{c}{$\gamma=15$} \\
\cmidrule(lr){2-3}
\cmidrule(lr){4-5}
\cmidrule(lr){6-7}
\cmidrule(lr){8-9}
& Speedup & $\tau$ & Speedup & $\tau$ & Speedup & $\tau$ & Speedup & $\tau$ \\
\midrule
Causal    & 8.29 &  9.81 & 8.50 & 10.00 & 8.40 &  9.99 & 8.41 &  9.96 \\
Diffusion & 5.46 &  6.45 & 8.16 &  9.65 & 8.36 &  9.72 & 6.17 &  7.19 \\
\bottomrule
\end{tabular}
\end{table}

Table~\ref{tab:ablation_gamma_arch} compares causal and diffusion heads under different choices of $\gamma$, the parameter that controls how aggressively the DFlash training objective downweights per-position loss at positions far from each anchor token. Specifically, position $i$ within a block contributes to the training loss with weight $w_i = \exp\!\left(-\max(i - i_{\mathrm{anchor}},\, 0)/\gamma\right)$, where $\gamma=0$ is interpreted as the limit of uniform per-position weighting (no decay). The causal head is robust across $\gamma$, while the diffusion head is sensitive to $\gamma$ and collapses at the endpoints (from $8.36\times$ at $\gamma=7$ to $5.46\times$ at $\gamma=0$ and $6.17\times$ at $\gamma=15$).




\paragraph{A failure example.}
On MATH-500 prompt $0$ at decode step $0$, the diffusion-head draft ($\gamma=0$) ranks ``\texttt{ given told that}'' first in its tree, with cumulative draft surrogate $\sum_i \log r_i = -3.76$ but cumulative target conditional $\sum_i \log p(y_i \mid x, \pi_{<v_i}) = -63.32$ nats: the branch combines two mutually exclusive openers, exemplifying the failure mode discussed in Section~\ref{par:why_causality}. The actually-coherent ``\texttt{ are given that the}'' (target joint $-0.08$) sits at rank~3. The causal-head draft on the same prompt ranks ``\texttt{ are told that}'' first, with surrogate ${\approx}$ target joint (gap $-0.34$); tree verification accepts $6$ tokens from the causal-head tree but only $4$ from the diffusion-head tree (visualized in Figure~\ref{fig:tree_quality_case_study} of Appendix~\ref{app:tree_quality_case_study}).

\paragraph{Aggregate generalization.}
Across MATH-500 prompts $0$--$49$, the diffusion-head's rank-$1$ gap exceeds the causal-head's on $92\%$ of prompts, with a median $5\times$ larger ($+62.81$ vs.\ $+12.36$ nats); the mean per-step accepted-token ratio is $1.95\times$, closely matching the macroscopic $1.52\times$ speedup ratio in Table~\ref{tab:ablation_gamma_arch}. Full branch-level tables and the rank-$1$ gap distribution are in Appendix~\ref{app:tree_quality_case_study}.

\subsection{Future Work}
Several directions remain for future work. On the serving side, dynamic policies could adapt the draft budget to system load, while further systems optimization could reduce tree-verification-specific overheads, for example by using separate CUDA streams to overlap drafting with KV-block compaction. More flexible tree construction could allocate budget across depth and width based on branch confidence. Future work could also further exploit the causal structure of the draft model, for example through explicit causal factorization among draft tokens. Finally, hybrid policies could switch between linear and tree drafting for tasks where branching offers limited benefit.
\section{Conclusion}

We presented \capsysname, a causal parallel tree drafting framework for speculative decoding that jointly addresses the two factors limiting end-to-end speedup: drafting latency and accepted length. By causally coupling draftrepresentations while generating all depth-wise candidate distributions in a single low-cost pass, JetSpec makes tree construction better aligned with the target model’s autoregressive factorization. Across dense and MoE Qwen3 models and vLLM serving settings, \sysname consistently turns larger draft budgets into higher wall-clock speedup, for MATH-500, reaching $9.64\times$ on Qwen3-8B, $9.45\times$ on Qwen3-30B-A3B, and up to 4.33$\times$ serving throughput speedup on a single H100. These results show that causal coupling across draft positions improves long-budget speculative drafting while retaining one-pass efficiency.


\newpage
\section*{AI Use Statement}

In this work, we used generative AI tools to provide feedback on technical formulations and qualitative data analysis, we also used AI to help with method implementations. We have not used generative AI tools to generate synthetic datasets, develop theoretical models, assist in writing proofs, propose hypotheses, translate content, clean or reformat datasets, or interpret experimental results. Additionally, we used generative AI tools to edit and refine manuscript text for clarity and readability, suggest improvements to presentation and structure, and assist with literature search and verification. We reviewed all AI-assisted work and manually verified technical claims, citations, and suggested revisions against the original sources and our own experiments. We take responsibility for the final content of this work, including text, claims, and artifacts produced with the aid of generative AI.

\section*{Reproducibility Statement}
We provide detailed descriptions of the proposed method, training procedure, and system implementation to facilitate reproducibility. Appendix~\ref{app:parallel_tree_drafting_algorithm} specifies the parallel tree-drafting algorithm and scoring procedure; Appendix~\ref{app:training_details} reports the draft-head architecture, target-feature extraction, training data construction, optimization settings, and loss configurations; Appendix~\ref{app:vllm_implementation_details} describes the vLLM integration and optimized tree-verification implementation. We also provide the experimental code in the supplementary material.

\section*{Ethics Statement}

All authors have read and adhere to the \href{https://iclr.cc/public/CodeOfEthics}{ICLR Code of Ethics}. This work does not involve human subjects, sensitive personal data, or experiments with the potential to cause harm. No confidential or proprietary data were used. The methods and experiments were conducted in accordance with principles of research integrity, fairness, and transparency. All conclusions are the sole responsibility of the authors.

\bibliography{ptd}
\bibliographystyle{iclr2027_conference}

\newpage
\appendix

\section{Tree Drafting Quality Analysis}
\label{app:tree_quality_case_study}

This appendix provides the full data behind the rank-$1$ case study and the $N=50$ aggregate statistics summarized in Section~\ref{sec:experiment_diffusion_tree_weakness}. Figure~\ref{fig:tree_quality_case_study} visualizes the rank-$1$ + rank-$3$ contrast for both heads on MATH-500 prompt $\#0$, decode step $0$ (root token \texttt{``We''}); the full top-$5$ branches and the rank-$1$ gap distribution follow.


\begin{figure}[th]
\centering
\begin{tikzpicture}[
    tok/.style={
        rectangle, draw, rounded corners=2pt, inner sep=4pt,
        font=\ttfamily, text height=1.6ex, text depth=0.4ex
    },
    faithful/.style={tok, fill=green!10, draw=green!55!black},
    incoherent/.style={tok, fill=red!8, draw=red!65!black},
    rowlabel/.style={font=\footnotesize, anchor=east, inner sep=2pt},
    panelhdr/.style={font=\bfseries, anchor=west},
    annot/.style={font=\footnotesize, anchor=west, inner sep=2pt}
]

\node[panelhdr] at (0, 4.6) {(a) Causal head, $\gamma=0$};
\node[rowlabel]                              at (1.7, 3.8)        {rank 1\, \textcolor{green!55!black}{$\checkmark$}};
\node[faithful, anchor=west]                  at (1.75, 3.8) (c1a) {are};
\node[faithful, anchor=west, xshift=1.5mm]    at (c1a.east)  (c1b) {told};
\node[faithful, anchor=west, xshift=1.5mm]    at (c1b.east)  (c1c) {that};
\node[annot, xshift=4mm]                      at (c1c.east)        {gap $= \mathbf{-0.34}$\quad\textcolor{green!55!black}{\itshape faithful}\quad$\Rightarrow$ verifier accepts \textbf{6} tokens};
\node[rowlabel]                              at (1.7, 2.9)        {rank 3\, \textcolor{red!65!black}{$\times$}};
\node[incoherent, anchor=west]                at (1.75, 2.9) (c3a) {are};
\node[incoherent, anchor=west, xshift=1.5mm]  at (c3a.east)  (c3b) {given};
\node[incoherent, anchor=west, xshift=1.5mm]  at (c3b.east)  (c3c) {that};
\node[incoherent, anchor=west, xshift=1.5mm]  at (c3c.east)  (c3d) {the2product};
\node[annot, xshift=4mm]                      at (c3d.east)        {gap $= +42.50$};

\node[panelhdr] at (0, 1.7) {(b) Diffusion head, $\gamma=0$};
\node[rowlabel]                              at (1.7, 0.9)        {rank 1\, \textcolor{red!65!black}{$\times$}};
\node[incoherent, anchor=west]                at (1.75, 0.9) (d1a) {given};
\node[incoherent, anchor=west, xshift=1.5mm]  at (d1a.east)  (d1b) {told};
\node[incoherent, anchor=west, xshift=1.5mm]  at (d1b.east)  (d1c) {that};
\node[annot, xshift=4mm]                      at (d1c.east)        {gap $= \mathbf{+59.56}$\quad\textcolor{red!65!black}{\itshape incoherent}\quad$\Rightarrow$ verifier accepts \textbf{4} tokens};
\node[rowlabel]                              at (1.7, 0.0)        {rank 3\, \textcolor{green!55!black}{$\checkmark$}};
\node[faithful, anchor=west]                  at (1.75, 0.0) (d3a) {are};
\node[faithful, anchor=west, xshift=1.5mm]    at (d3a.east)  (d3b) {given};
\node[faithful, anchor=west, xshift=1.5mm]    at (d3b.east)  (d3c) {that};
\node[faithful, anchor=west, xshift=1.5mm]    at (d3c.east)  (d3d) {the};
\node[annot, xshift=4mm]                      at (d3d.east)        {gap $= -3.69$\quad\textcolor{green!55!black}{\itshape faithful, but at rank 3}};

\end{tikzpicture}

\caption{\textbf{Tree-quality failure mode at MATH-500 prompt $\#0$, decode step $0$.}
Both heads draft from the same prefix (last token \texttt{``We''}). The causal head's rank-$1$ branch (``\texttt{ are told that}'') is faithful: target joint $\Sigma\log p \approx \Sigma\log r$, so tree verification walks $6$ tokens along it. The diffusion head's rank-$1$ branch (``\texttt{ given told that}'') is incoherent (target joint $\Sigma\log p = -63.32$ nats, i.e.\ probability ${\approx}e^{-63}$) because its branch-agnostic per-position predictor composes ``\texttt{ given}'' (depth $1$) and ``\texttt{ told}'' (depth $2$) independently in the surrogate, even though no real continuation places these two words consecutively. The actually-coherent ``\texttt{ are given that the}'' is in the diffusion tree but only at rank $3$; the surrogate fails to promote it, and the verifier accepts only $4$ tokens.}

\label{fig:tree_quality_case_study}
\end{figure}

\subsection{Setup}
\label{app:tree_quality_setup}

We modify the speculative decoding loop to dump the constructed tree at a chosen decode step. For every node $v$ we record the cumulative draft surrogate $\sum_{i=1}^{\mathrm{depth}(v)} \log r_i(y_{v_i}\mid x)$ and the cumulative target conditional $\sum_{i=1}^{\mathrm{depth}(v)} \log p(y_{v_i}\mid x, \pi_{<v_i})$, along with the tree-verifier's accepted path. The two drafts compared are a causal head and a diffusion head, both trained at LR $3{\times}10^{-4}$ with $\gamma=0$, evaluated on $4{\times}$B200 with block size $N=16$ (also the maximum draft depth), FlashAttention-2, and Triton tree-attention. Both heads share the prefill and produce $N$ parallel position-wise distributions $r_1, \ldots, r_N$; tree construction uses the cumulative-log-probability heuristic with branching width $W=7$ and node budget $B=255$.

\subsection{Top-5 Branches at the Canonical Example}
\label{app:tree_quality_top5}


Table~\ref{tab:tree_case_study_full} extends Figure~\ref{fig:tree_quality_case_study} to the full top-$5$ branches of each head's tree at MATH-500 prompt $\#0$, decode step $0$ (root token \texttt{``We''}). The pattern reported in the main text repeats throughout the tree. For the diffusion head, top-$2$ and top-$4$ both combine ``\texttt{ given}'' at depth~$1$ and ``\texttt{ told}'' at depth~$2$ with target joints below $-50$ nats; only rank-$3$ (target joint $-0.08$) is coherent. For the causal head, the rank-$1$ branch (``\texttt{ are told that}'') is faithful (gap $-0.34$); ranks $2$--$5$ all share the off-argmax depth-$2$ token ``\texttt{ given}'' (vs.\ rank-$1$'s argmax ``\texttt{ told}'') and degrade at depths $4$--$5$ (e.g.\ ``\texttt{the2product}'' at rank $3$, gap $+42.50$), consistent with the off-argmax inheritance described in \S\ref{app:tree_quality_mechanism}: each branch's depth-$d$ marginal was anchored to the argmax extension at depth $d-1$, so off-argmax branches inherit a conditioning context that does not match their own ancestor token.

\begin{table*}[th]
\centering
\small
\caption{Top-5 branches of each head's tree at MATH-500 prompt $0$, decode step $0$. Both drafts run on the identical prefix with the same target. $\Sigma\log r$ and $\Sigma\log p$ in nats; $\Delta = $ surrogate $-$ target.}
\label{tab:tree_case_study_full}
\setlength{\tabcolsep}{4.5pt}
\begin{tabular}{lllrrr}
\toprule
Architecture & Rank & Branch text & $\Sigma\log r$ & $\Sigma\log p$ & $\Delta$ \\
\midrule
\multirow{5}{*}{Causal $\gamma{=}0$}
 & 1 & \texttt{"~are~told~that"}                & $-3.88$ & $-3.54$  & $-0.34$  \\
 & 2 & \texttt{"~are~given~that~the~product~**"} & $-3.96$ & $-11.68$ & $+7.73$  \\
 & 3 & \texttt{"~are~given~that~the2product"}    & $-3.96$ & $-46.46$ & $+42.50$ \\
 & 4 & \texttt{"~are~given~that~the2~of"}        & $-3.96$ & $-49.71$ & $+45.75$ \\
 & 5 & \texttt{"~are~given:\textbackslash n\textbackslash n~the~**"} & $-4.01$ & $-40.59$ & $+36.58$ \\
\midrule
\multirow{5}{*}{Diffusion $\gamma{=}0$}
 & 1 & \texttt{"~given~told~that~**"}            & $-3.76$ & $-63.32$ & $+59.56$ \\
 & 2 & \texttt{"~given~told:\textbackslash n\textbackslash n~the"} & $-3.77$ & $-90.78$ & $+87.02$ \\
 & 3 & \texttt{"~are~given~that~the"}            & $-3.77$ & $-0.08$  & $-3.69$  \\
 & 4 & \texttt{"~told~given~that"}               & $-3.84$ & $-50.94$ & $+47.10$ \\
 & 5 & \texttt{"~given~told~that:\textbackslash n\textbackslash n~product"} & $-3.87$ & $-96.44$ & $+92.57$ \\
\bottomrule
\end{tabular}
\end{table*}

\subsection{Gap Distribution over $50$ Prompts}
\label{app:tree_quality_distribution}


Table~\ref{tab:tree_case_aggregate} reports the rank-$1$ gap distribution across MATH-500 prompts $0$--$49$ for both heads at $\gamma=0$ and at $\gamma=7$ (DFlash's best macroscopic loss-weighting setting, Table~\ref{tab:ablation_gamma_arch}). At $\gamma=0$, the diffusion head exhibits a dramatic upper-tail failure: $26\%$ of prompts have rank-$1$ gap $\geq +80$ nats (joint target probability $<10^{-35}$), versus $0\%$ for the causal head; faithful rank-$1$ (gap $<+5$ nats) appears in only $6\%$ of diffusion prompts versus $42\%$ for causal. The macroscopic effect is direct: mean accepted length is $4.84$ tokens for diffusion vs $9.46$ for causal at $\gamma=0$. At $\gamma=7$, the diffusion head's tree quality recovers substantially — its extreme tail drops to $4\%$ and faithful rank-$1$ rises to $26\%$ — with mean acceptance reaching $9.42$ tokens. It does not, however, match the causal head, which retains the lower extreme tail ($2\%$) and higher mean acceptance ($9.64$) at $\gamma=7$. The takeaway is that the causal head's rank-$1$ faithfulness is robust across $\gamma$, whereas the diffusion head depends on loss-weighting tuning to recover from the $\gamma=0$ failure mode. \capsysnamespace's contribution is the structural robustness this provides: the causal mask between depths obviates the need to tune $\gamma$.


\begin{table*}[th]
\centering
\small
\renewcommand{\arraystretch}{1.15}
\caption{Distribution of rank-$1$ surrogate--target gaps across MATH-500 prompts $0$--$49$ (each at the prompt's natural step-$0$ tree). $\gamma=7$ is DFlash's best macroscopic loss-weighting setting (Table~\ref{tab:ablation_gamma_arch}). Mean accepted length per step is reported in the bottom row for context.}
\label{tab:tree_case_aggregate}
\setlength{\tabcolsep}{4.5pt}
\begin{tabular}{lcccc}
\toprule
\multirow{2}{*}{Rank-$1$ gap (nats)} & \multicolumn{2}{c}{$\gamma=0$} & \multicolumn{2}{c}{$\gamma=7$} \\
\cmidrule(lr){2-3}\cmidrule(lr){4-5}
 & Causal & Diffusion & Causal & Diffusion \\
\midrule
$<+5$ (faithful)     & $42\%$ ($21/50$) & $6\%$  ($3/50$)  & $32\%$ ($16/50$) & $26\%$ ($13/50$) \\
$[+5, +20)$          & $12\%$ ($6/50$)  & $0\%$  ($0/50$)  & $6\%$  ($3/50$)  & $14\%$ ($7/50$)  \\
$[+20, +40)$         & $24\%$ ($12/50$) & $4\%$  ($2/50$)  & $32\%$ ($16/50$) & $20\%$ ($10/50$) \\
$[+40, +80)$         & $22\%$ ($11/50$) & $64\%$ ($32/50$) & $28\%$ ($14/50$) & $36\%$ ($18/50$) \\
$\geq +80$ (extreme) & $0\%$  ($0/50$)  & $26\%$ ($13/50$) & $2\%$  ($1/50$)  & $4\%$  ($2/50$)  \\
\midrule
Median gap (nats)            & $+12.36$ & $+62.81$ & $+27.87$ & $+32.69$ \\
Mean gap (nats)              & $+20.46$ & $+63.25$ & $+26.58$ & $+32.07$ \\
Mean accepted length         & $9.46$   & $4.84$   & $9.64$   & $9.42$   \\
\bottomrule
\end{tabular}
\end{table*}

\subsection{Why the Failure Mode Is Structural}
\label{app:tree_quality_mechanism}

The diffusion head produces all $N$ position-wise distributions $r_1, \ldots, r_N$ from a single shared hidden state with no causal mask between depths, so $r_2$ does not condition on $y_1$. When $r_1$ and $r_2$ both place high mass on tokens that fit a given depth marginally but cannot follow each other syntactically (e.g.\ ``\texttt{ given}'' at depth~$1$ and ``\texttt{ told}'' at depth~$2$), the surrogate $r_1(y_1)\cdot r_2(y_2)$ promotes their composition into the rank-$1$ branch. The causal head's parallel marginals are produced with causal masking between depths, so $r_2$ is anchored to the model's own argmax at depth~$1$; off-argmax branches in the heap still inherit this anchored $r_2$, but the rank-$1$ branch (the argmax-aligned one, which dominates verifier behavior) receives a faithful surrogate score. This anchoring is what closes the gap between $q_{\mathrm{sur}}$ and the target's branch-conditional joint at the head of the tree, and it is what allows the causal head to remain robust across $\gamma$: the rank-$1$ alignment does not depend on a loss-weighting prior over depths.


\section{Tree-Drafting Algorithm Details}
\label{app:parallel_tree_drafting_algorithm}

Algorithm~\ref{alg:tree_construction} describes how \capsysnamespace constructs a candidate tree under a fixed node budget. Starting from the root prefix \(x\), the algorithm maintains a priority queue of expandable nodes ranked by a branch scoring function \(\mathrm{Score}(\cdot)\). At each iteration, it expands the highest-scoring node by adding up to \(W\) children at the next depth, unless the node has already reached the maximum draft depth \(N\). Each new child is inserted back into the queue with its updated path score, and the process continues until the total node budget \(B\) is exhausted. The returned tree \(\mathcal{T}(x)\) therefore contains the highest-scoring candidate branches according to the chosen scoring rule.

\begin{algorithm}[ht]
\caption{Parallel Tree Drafting}
\label{alg:tree_construction}
\begin{algorithmic}[1]
\Require Prefix \(x\), maximum draft depth \(N\), branching width \(W\), node budget \(B\), scoring function \(\mathrm{Score}(\cdot)\)
\Ensure Candidate tree \(\mathcal{T}(x)\)

\State Initialize tree \(\mathcal{T}\) with root node \(v_0\)
\State Initialize priority queue \(\mathcal{Q}\leftarrow\{(v_0, \mathrm{Score}(\pi(v_0)))\}\)

\While{\(|\mathcal{V}_{\mathcal{T}}| < B\) and \(\mathcal{Q}\neq\emptyset\)}
    \State Pop the highest-scoring node \(v\) from \(\mathcal{Q}\)
    \If{\(\mathrm{depth}(v) = N\)}
        \State \textbf{continue}
    \EndIf
    \State Obtain up to \(W\) candidate children \(\mathcal{C}(v)\) for the next depth
    \For{\(y \in \mathcal{C}(v)\)}
        \If{\(|\mathcal{V}_{\mathcal{T}}| = B\)}
            \State \textbf{break}
        \EndIf
        \State Add child node \(u\) with token \(y_u=y\) and parent \(v\) to \(\mathcal{T}\)
        \State Compute \(s_u \leftarrow \mathrm{Score}(\pi(u))\)
        \State Push \((u,s_u)\) into \(\mathcal{Q}\)
    \EndFor
\EndWhile

\State \Return \(\mathcal{T}(x)=\{\pi(v)\mid v\in\mathcal{V}_{\mathcal{T}}\}\)
\end{algorithmic}
\end{algorithm}

\section{Tree-Drafting Algorithm Ablation}
\label{app:tree_draft_ablation}


We compare three tree-construction scoring algorithms at the production setting: cumulative draft log-probability (\texttt{accum\_logp}, our default), entropy-guided (per-depth entropy alone), and hybrid (cumulative log-probability plus $\alpha$-weighted per-depth entropy). Results are summarized in Table~\ref{tab:ablation_tree_draft}.

\paragraph{Cumulative log-probability scoring.}
\texttt{accum\_logp} achieves $8.15\times$ speedup (average accepted length $\tau = 9.81$) and is our production default. It expands the tree by best-first heap priority on cumulative draft log-probability along each branch, prioritizing high-likelihood continuations under the budget.

\paragraph{Entropy-only collapse.}
Entropy-guided scoring (frontier priority by per-depth marginal entropy alone, with no cumulative-log-prob signal) drops to $4.76\times$ ($-42\%$ relative to \texttt{accum\_logp}), confirming that per-depth entropy alone is insufficient to identify high-acceptance branches.

\paragraph{Hybrid scoring with increasing $\alpha$.}
Hybrid scoring uses $\sum_i \log r_i + \alpha \cdot H_i$, with per-depth entropy $H_i$. At $\alpha \leq 1$, hybrid converges within ${\sim}1.5\%$ of \texttt{accum\_logp} ($8.15$--$8.27\times$); increasing $\alpha$ monotonically degrades acceptance, reaching $7.42\times$ at $\alpha{=}8$. The $\alpha\to 0$ limit reduces to pure cumulative log-probability (consistent with \texttt{accum\_logp} sitting on the same plateau as low-$\alpha$ hybrid); the $\alpha\to\infty$ limit corresponds to entropy-only scoring (consistent with the $4.76\times$ collapse).

\begin{table*}[th]
\centering
\small
\caption{Tree-construction algorithm ablation on MATH-500 ($n=500$) with \sysname at the production setting (causal head, LR $3{\times}10^{-4}$, Forward-KL distillation, $\gamma=0$). Hybrid scoring is $\sum_i \log r_i + \alpha \cdot H_i$ with per-depth entropy $H_i$. We report speedup and average accepted length $\tau$.}
\label{tab:ablation_tree_draft}
\setlength{\tabcolsep}{4.5pt}
\begin{tabular}{lcc}
\toprule
& Speedup & $\tau$ \\
\midrule
\multicolumn{3}{l}{\emph{Algorithm}} \\
\quad \textbf{Accum.\ log-prob (default)} & \textbf{8.15} & 9.81 \\
\quad Entropy-guided            & 4.76 & 5.52 \\
\midrule
\multicolumn{3}{l}{\emph{Hybrid (log-prob $+$ $\alpha\cdot$ entropy)}} \\
\quad $\alpha = 0.25$           & 8.27 & 9.81 \\
\quad $\alpha = 0.5$            & 8.22 & 9.78 \\
\quad $\alpha = 1.0$            & 8.15 & 9.78 \\
\quad $\alpha = 2.0$            & 7.98 & 9.63 \\
\quad $\alpha = 4.0$            & 7.75 & 9.27 \\
\quad $\alpha = 8.0$            & 7.42 & 9.00 \\
\bottomrule
\end{tabular}
\end{table*}


\section{Training Details}
\label{app:training_details}

For both DFlash and \sysname, we perform a learning-rate sweep from $1{\times}10^{-4}$ to $1{\times}10^{-3}$ with five settings. We find that $3{\times}10^{-4}$ and $6{\times}10^{-4}$ generally perform best, with different tasks favoring different choices. Since their overall performance is comparable, we report results with $3{\times}10^{-4}$ by default. All draft-head training runs are conducted on 8 H100 GPUs with a micro batch of 2. 

\begin{figure}[h]
    \centering
    \includegraphics[width=0.92\linewidth]{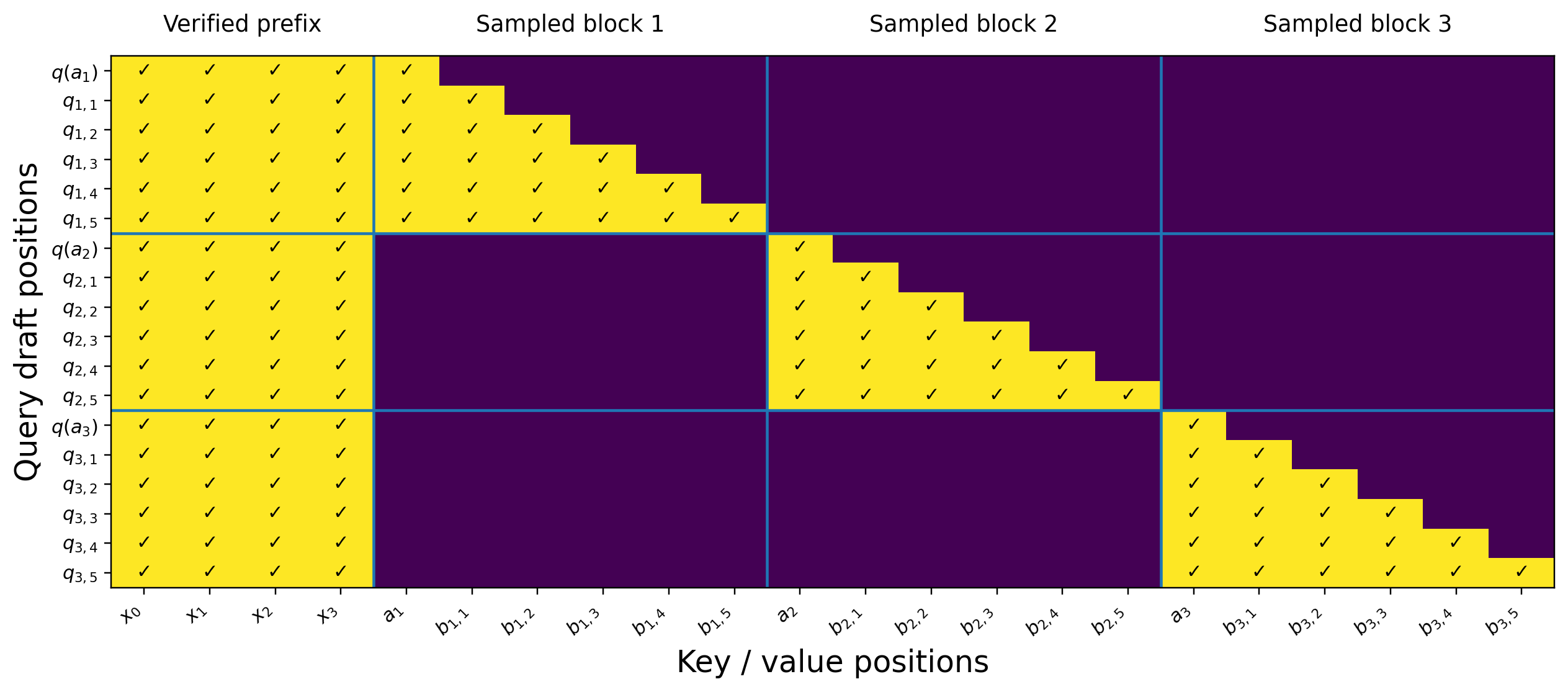}
    \caption{
    Causal attention mask used for training with multiple sampled blocks. Each query can attend to the full verified prefix and to the anchor plus earlier positions within its own block, but cannot attend to future positions or positions from other sampled blocks.
    }
    \label{fig:training_attention_mask}
\end{figure}

For a fair comparison with DFlash, the \capsysnamespace draft head is trained with block size 16, corresponding to a maximum tree depth of 16, and uses fused hidden features extracted from the frozen target model. During training, each block keeps the first token as the anchor and replaces the remaining positions with mask-token embeddings. The causal head predicts all masked future tokens in parallel, while each position can attend only to the prefix and earlier positions within the same block, ensuring that the draft distribution follows the autoregressive order. We sample random anchor positions from each sequence, use up to 512 anchors per example. An example of the attention mask with 3 anchors for 3 blocks is provided in Figure~\ref{fig:training_attention_mask}.

\capsysnamespace reuses intermediate representations from the frozen target model as draft-head context. For Qwen3-8B, we extract hidden states from target layers $\{1,9,17,25,33\}$ out of the 36-layer target model, concatenate them along the channel dimension, and project the resulting $5d$ feature back to hidden size $d=4096$ through a bias-free linear layer followed by RMSNorm. The draft head is implemented as a lightweight Qwen3-style decoder with 5 layers, 32 attention heads, 8 KV heads, head dimension 128, and MLP intermediate size 12288. In each draft layer, the projected target feature is injected as contextual key/value states and concatenated with the draft-token hidden states, allowing the causal draft head to condition on rich target-model features while keeping the target model frozen.

For ablations, we consider DFlash-style exponential depth weighting. Since depth weighting can implicitly bias the bidirectional DFlash head toward left-to-right prediction, in main results reported in Table~\ref{tab:main_results_linear_budget_sweep} and Table~\ref{tab:main_results_tree_budget_sweep}, we remove it when isolating the effect of causal masking versus the native block-diffusion head design. For distillation runs, teacher logits are obtained from the frozen target model and aligned with the student prediction positions; the draft head is trained with a temperature-scaled soft-label distillation loss.

\begin{figure}[t]
    \centering
    \includegraphics[width=0.88\linewidth]{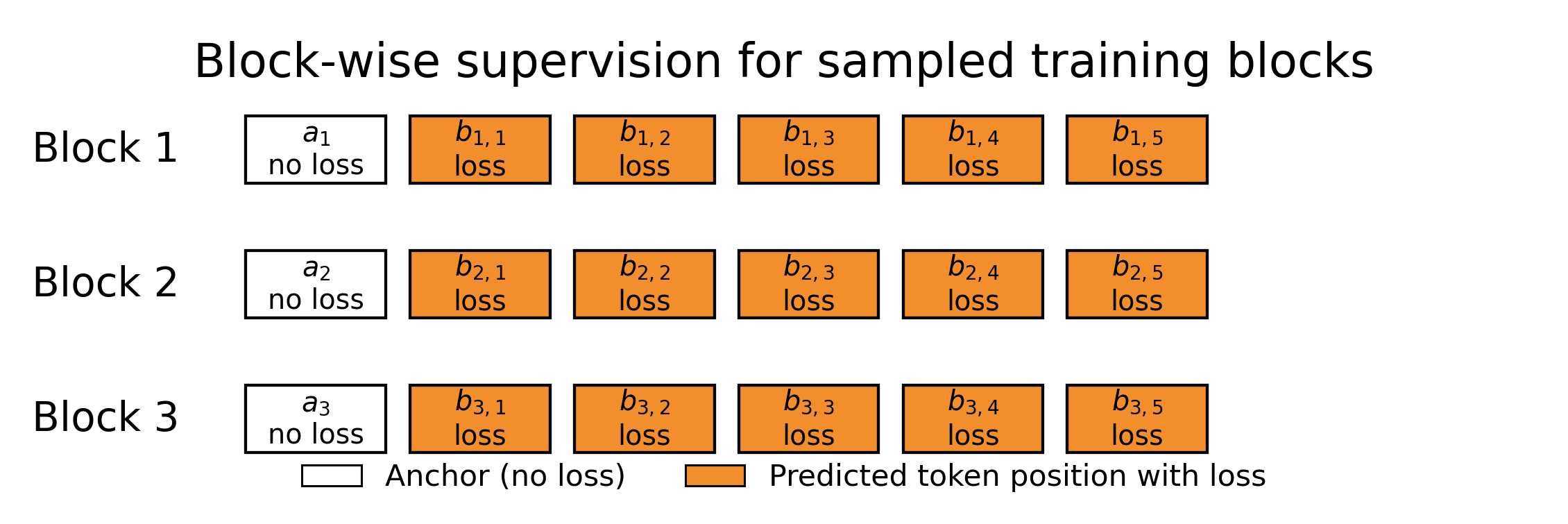}
    \caption{
    Each sampled block includes an anchor position and multiple future token positions.
    The anchor is retained as block context and excluded from the loss, while loss is applied only to future token positions within each block.
    }
    \label{fig:training_loss_mask}
\end{figure}



\section{vLLM Implementation Details}
\label{app:vllm_implementation_details}

We implement causal parallel tree drafting inside vLLM. The draft head uses causal attention, while its candidates are organized as a speculative tree rather than a linear block.

Given the target hidden states at the current decoding step, the proposer produces draft logits for a block of future positions. For each position, it extracts the top-\(k\) tokens and log-probabilities, where \(k\) is the tree width, and constructs a speculative tree under a fixed node budget. Our implementation supports several expansion scores, including cumulative log-probability, entropy-guided scoring, and hybrid scoring, as discussed in Appendix~\ref{app:tree_draft_ablation}. We use cumulative log-probability in our experiments. The resulting tree stores token ids, parent indices, and node depths, with the root corresponding to the already-sampled next token.

During verification, the tree metadata is passed to the target model as parent indices and depths for all speculative nodes. The target verifies all tree nodes in one forward pass using a tree attention mask. To reduce verification overhead, we implement an optimized fused paged tree-attention kernel: it builds the ancestor relation for speculative nodes, applies the tree mask inside attention, and verifies all candidates without materializing a dense per-request mask. Acceptance is computed by comparing each node token with the target model's greedy prediction at its parent, then selecting the deepest fully accepted root-to-node path. The accepted path is committed, and the target prediction at the final accepted node is used as the correction token.

Table~\ref{tab:vllm_batch_budget_sweep} reports an example end-to-end vLLM run on MATH-500 with a single H100 GPU. The results show how tree budget interacts with serving batch size: larger budgets improve low-batch latency by reducing verification rounds, while their relative gain decreases at larger batch sizes as verification and memory pressure become more prominent.

\begin{table}[ht]
\centering
\small
\caption{
vLLM serving performance of \capsysnamespace on Math-500 with Qwen3-8B on a single H100 GPU, evaluated across batch sizes and tree budgets (in parentheses). Each setting reports throughput in tokens/s (TPS), with speedup over AR decoding shown in parentheses.
}
\label{tab:vllm_batch_budget_sweep}
\begin{adjustbox}{max width=\textwidth}
\begin{tabular}{cccccc}
\toprule
Batch Size
& AR
& \sysname (16)
& \sysname (32)
& \sysname (64)
& \sysname (128) \\
\midrule
1  & 127.8 & 224.0 ($1.75 \times$) & 312.0 ($2.44 \times$) & 447.3 ($3.50 \times$) & 553.3 ($4.33 \times$) \\
2  & 163.3 & 266.2 ($1.63 \times$) & 360.3 ($2.21 \times$) & 516.0 ($3.16 \times$) & 621.6 ($3.81 \times$) \\
4  & 203.8 & 433.6 ($2.13 \times$) & 534.2 ($2.62 \times$) & 664.2 ($3.26 \times$) & 742.9 ($3.64 \times$) \\
8  & 246.2 & 679.3 ($2.76 \times$) & 839.3 ($3.41 \times$) & 859.3 ($3.49 \times$) & 803.5 ($3.26 \times$) \\
16 & 287.3 & 891.8 ($3.10 \times$) & 1094.6 ($3.81 \times$) & 995.8 ($3.47 \times$) & 803.1 ($2.80 \times$) \\
\bottomrule
\end{tabular}
\end{adjustbox}
\end{table}

\section{Ablation Study Details}
\label{app:ablation_details}

We provide detailed ablation results for the training and generalization settings summarized in Section~\ref{sec:experiments}. Unless otherwise specified, ablations are conducted on the Nemotron Post-Training Dataset V2 math split, using Qwen3-8B with block size 16, tree node budget 255, accumulated draft log-probability for tree construction, and the same evaluation protocol as the main experiments.

\paragraph{Learning Rate.}
Table~\ref{tab:ablation_lr} studies the effect of draft-head learning rate. Very small learning rates underfit the drafter, while performance plateaus around $3\times10^{-4}$; larger rates remain competitive but do not provide consistent additional gains.

\paragraph{Training Objective.}
Table~\ref{tab:ablation_loss} compares SFT, forward-KL distillation, and reverse-KL distillation. SFT and forward-KL perform similarly, while reverse-KL substantially degrades performance, suggesting that mode-seeking distillation is poorly matched to budgeted tree drafting, where preserving multiple plausible continuations is important.

\paragraph{Model Generalizability.}
Table~\ref{tab:ablation_model_generalizability} evaluates whether \sysname generalizes beyond dense Qwen3-8B. On the Qwen3-30B-A3B MoE target model, \sysname continues to outperform DDTree under the same training recipe, showing that causal parallel tree drafting is not specific to a single model architecture.

\paragraph{Training Data.}
Table~\ref{tab:ablation_training_data} compares regenerated target-model sequences with direct corpus training. Regenerated data gives the strongest performance by better matching the target model's own generation distribution. Corpus-trained \sysname is weaker but still effective.

\section{Empirical Per-token Drafting Cost on Modern Hardware}
\label{app:ptd-ptv}

Figure~\ref{fig:sd_speedup_vs_gamma} shows that speculative decoding scales better
with draft length when the per-token drafting cost is low. In this appendix, we
measure this cost for a practical DFlash-style draft head on a single H200 NVL
GPU. We use $N$ to denote the draft depth, i.e., the number of draft tokens
proposed in one block. We sweep context length and draft depth with request
batch size fixed to one, and report the per-draft-token cost ratio corresponding
to the coefficient $c$ in Eq.~\eqref{eq:sd_speedup}.

\paragraph{Setup.}
We profile a DFlash configuration using Qwen3-8B as the target model and
\texttt{z-lab/Qwen3-8B-DFlash-b16} as the draft head. Each cell reports
steady-state latency after 10 warmup steps and 50 measurement steps, measured
with CUDA events using the PyTorch SDPA backend. For context length $L$ and
draft depth $N$, let $T_{\mathrm{draft}}(N,L)$ denote the latency of one parallel
draft-head forward pass that proposes $N$ tokens, and let
$T_{\mathrm{verify}}(N,L)$ denote the latency of one parallel target-model
verification pass over the same $N$ candidates. Following Eq.~\eqref{eq:sd_speedup},
we define the per-draft-token cost coefficient as
\[
  c(N,L)
  =
  \frac{T_{\mathrm{draft}}(N,L)/N}
       {T_{\mathrm{verify}}(N,L)}
  =
  \frac{T_{\mathrm{draft}}(N,L)}
       {N\,T_{\mathrm{verify}}(N,L)}.
\]
The factor $1/N$ amortizes the parallel draft-head forward cost across the
$N$ proposed tokens. Lower $c$ indicates cheaper marginal drafting relative to
one target verification pass.

\paragraph{Results.}
Table~\ref{tab:c-per-token} reports $c$ in percent over a sweep of context
lengths $L\!\in\!\{128,\dots,4096\}$ and draft depths
$N\!\in\!\{1,2,4,8,16,32,64,128,256,512\}$.

\begin{table}[h]
    \centering
    \caption{Per-draft-token drafting cost ratio
    $c = T_{\mathrm{draft}}/(N\,T_{\mathrm{verify}})$ (\%) on a single H200
    NVL GPU, sweeping context length $L$ and draft depth $N$. This is the
    cost coefficient used in Eq.~\eqref{eq:sd_speedup} and
    Fig.~\ref{fig:sd_speedup_vs_gamma}. Lower is better.}
    \label{tab:c-per-token}
    \small
    \begin{tabular}{r|rrrrrrrrrr}
        \toprule
        $L \backslash N$ & $1$ & $2$ & $4$ & $8$ & $16$ & $32$ & $64$ & $128$ & $256$ & $512$ \\
        \midrule
        128 & 15.098 & 6.810 & 3.396 & 1.683 & 0.846 & 0.422 & 0.211 & 0.106 & 0.053 & 0.034 \\
        256 & 15.042 & 6.735 & 3.329 & 1.663 & 0.837 & 0.419 & 0.210 & 0.105 & 0.052 & 0.034 \\
        512 & 15.083 & 6.710 & 3.327 & 1.679 & 0.843 & 0.421 & 0.208 & 0.105 & 0.052 & 0.034 \\
        1024 & 15.029 & 6.720 & 3.327 & 1.690 & 0.845 & 0.420 & 0.209 & 0.104 & 0.054 & 0.035 \\
        2048 & 15.031 & 6.664 & 3.347 & 1.668 & 0.831 & 0.415 & 0.211 & 0.110 & 0.064 & 0.036 \\
        4096 & 18.295 & 8.796 & 4.411 & 2.233 & 1.129 & 0.566 & 0.289 & 0.146 & 0.073 & 0.037 \\
        \bottomrule
    \end{tabular}
\end{table}

Across the practically relevant regime ($L\!\le\!2048$, $N\!\ge\!16$), the
per-draft-token cost is below $1\%$ of one target verification pass and decreases
roughly as $1/N$, since the parallel draft-head cost is amortized across more
proposed tokens. At $L\!=\!1024$, for instance, $c$ drops from $0.845\%$ at
$N\!=\!16$ to $0.054\%$ at $N\!=\!256$, close to the ultra-low-cost regime in
Fig.~\ref{fig:sd_speedup_vs_gamma}. Longer contexts increase the cost mildly,
but $c$ remains below $0.1\%$ for $N\!\ge\!256$.

\paragraph{Implication.}
These measurements support the main scaling argument: increasing the draft depth
$N$ amortizes the draft-head cost and lowers the effective per-token cost
coefficient $c$. On the same hardware, $c$ decreases from $\approx6.7\%$ at
$N\!=\!2$ to $\approx0.05\%$ at $N\!=\!256$, moving from the typical-cost regime
toward the ultra-low-cost regime in Fig.~\ref{fig:sd_speedup_vs_gamma}. However,
as Fig.~\ref{fig:sd_speedup_vs_gamma} illustrates, low cost enables long-draft
scaling only when acceptance remains sufficiently high. This motivates causal
tree drafting as a way to improve acceptance quality in the low-cost speculative decoding.


\end{document}